\documentclass[letterpaper]{article} 
\usepackage[draft]{aaai2026}
\usepackage{times}  
\usepackage{helvet}  
\usepackage{courier}  
\usepackage[hyphens]{url}  
\usepackage{graphicx} 
\usepackage{natbib}  
\usepackage{caption} 
\usepackage{algorithm}
\usepackage{algorithmic}

\usepackage{newfloat}
\usepackage{listings}
\DeclareCaptionStyle{ruled}{labelfont=normalfont,labelsep=colon,strut=off} 
\floatstyle{ruled}
\newfloat{listing}{tb}{lst}{}
\floatname{listing}{Listing}
\title{The Strong Lottery Ticket Hypothesis for Multi-Head Attention Mechanisms}

\author {
    Hikari Otsuka\textsuperscript{\rm{1}, \textdagger},
    Daiki Chijiwa\textsuperscript{\rm 2},
    Yasuyuki Okoshi\textsuperscript{\rm 1},\\
    Daichi Fujiki\textsuperscript{\rm 1},
    Susumu Takeuchi\textsuperscript{\rm 2},
    Masato Motomura\textsuperscript{\rm 1}
}
\affiliations {
    \textsuperscript{\rm 1}Institute of Science Tokyo\quad
    \textsuperscript{\rm 2}NTT, Inc.\\
    \textsuperscript{\textdagger}otsuka.hikari@artic.iir.isct.ac.jp
}

\usepackage{bibentry}

\usepackage{mathtools}
\usepackage{amsthm}
\usepackage{thmtools}
\usepackage{cleveref}
\usepackage{amssymb}
\usepackage{amsthm}

\newtheorem{theorem}{Theorem}
\newtheorem{lemma}[theorem]{Lemma}
\theoremstyle{definition}

\usepackage{subcaption}
\usepackage{macros}
\def\bm#1{\boldsymbol{#1}}
\allowdisplaybreaks

\begin{document}

\maketitle

\begin{abstract}
    The strong lottery ticket hypothesis (SLTH) conjectures that high-performing subnetworks, called strong lottery tickets (SLTs), are hidden in randomly initialized neural networks.
Although recent theoretical studies have established the SLTH across various neural architectures, the SLTH for transformer architectures still lacks theoretical understanding.
In particular, the current theory of the SLTH does not yet account for the multi-head attention (MHA) mechanism, a core component of transformers.
To address this gap, we introduce a theoretical analysis of the existence of SLTs within MHAs.
We prove that, if a randomly initialized MHA of $H$ heads and input dimension $d$ has the hidden dimension $O(d\log(Hd^{3/2}))$ for the key and value, it contains an SLT that approximates an arbitrary MHA with the same input dimension with high probability.
Furthermore, by leveraging this theory for MHAs, we extend the SLTH to transformers without normalization layers.
We empirically validate our theoretical findings, demonstrating that the approximation error between the SLT within a source model (MHA and transformer) and an approximate target counterpart decreases exponentially by increasing the hidden dimension of the source model.

\end{abstract}


\begin{figure}[tb]
    \centering
    \includegraphics[width=0.94\columnwidth]{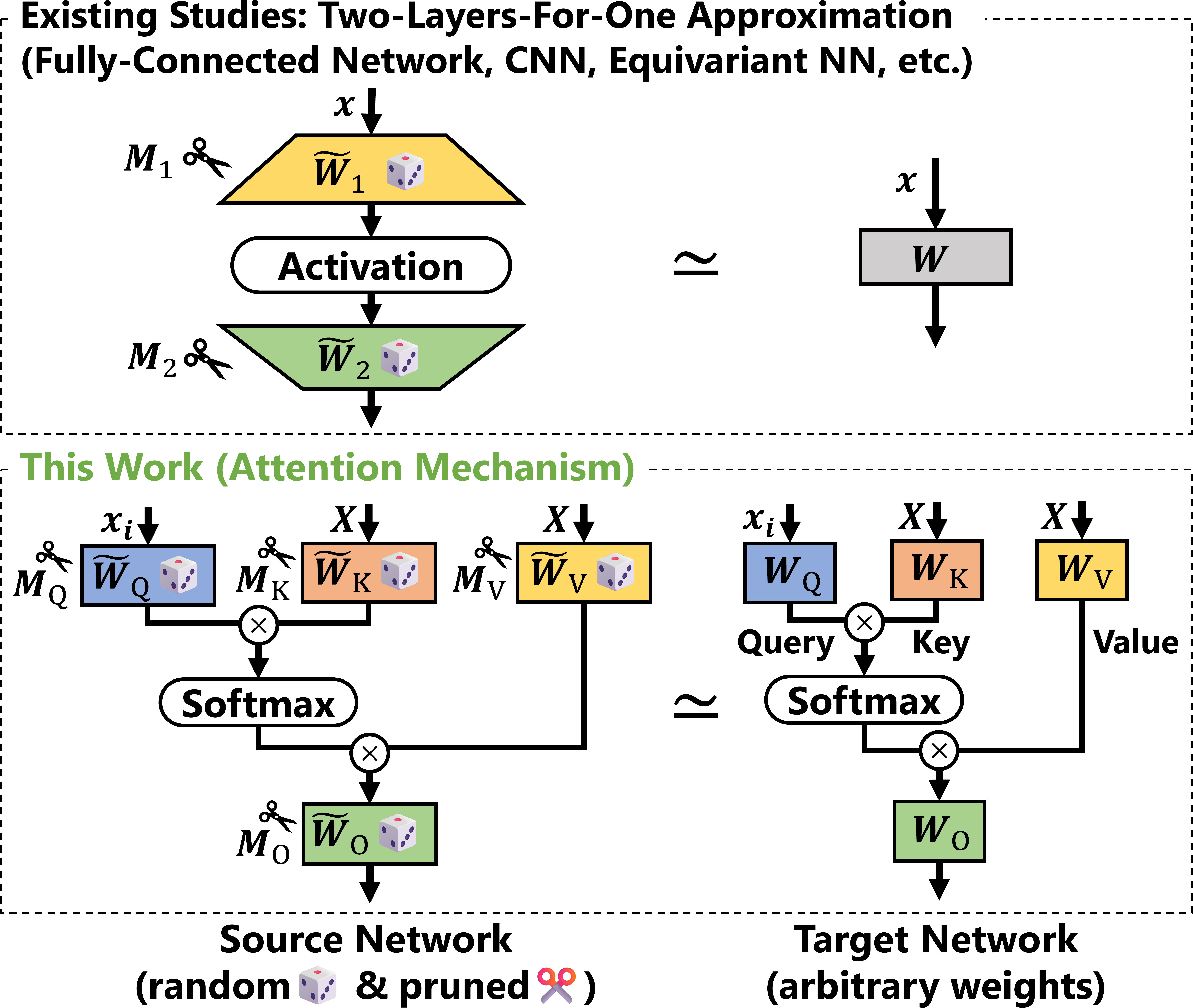}
    \caption{
        Comparison of the approximation techniques in conventional theories of the SLTH (top) and in our attention-specific approach (bottom).
        This work demonstrates that an arbitrary attention mechanism can be approximated by pruning a randomly initialized one.
        }
    \label{fig:outline}
\end{figure}
    
\section{Introduction}
    The \textit{lottery ticket hypothesis}~\citep{frankle2018the}---overparameterized networks contain subnetworks that achieve comparable accuracy to fully trained networks even if trained in isolation---presented new possibilities for compact and high-performing models inherent in recent deep neural networks.
    Later, a stronger claim, which is formally defined as the \textit{strong lottery ticket hypothesis} (SLTH), was proposed~\citep{ramanujan2020s, malach2020proving}: overparameterized networks contain subnetworks (called \textit{strong lottery tickets} (SLTs)) that achieve comparable accuracy to the trained dense network even without any training.
    Whether such subnetworks exist is a fascinating question in itself, and studying them can bring us closer to understanding the principles behind overparameterized models.

    The rigorous proof for the SLTH was firstly established in fully-connected networks.
    Early studies showed that a randomly-weighted fully-connected network of sufficient width (a \textit{source network}) contains an SLT, which approximates an arbitrary fully-connected network with half the depth (a \textit{target network})~\citep{malach2020proving, orseau2020logarithmic, pensia2020optimal}.
    These theories are built on the foundational argument called a \textit{two-layers-for-one approximation}: a two-layer source network with random weight matrices contains an SLT that approximates a single-layer target network with an arbitrary weight matrix (the top panel of \Cref{fig:outline}).
    Following this finding, subsequent studies have succeeded in proving the existence of SLTs in more complex networks, such as convolutional and equivariant networks~\citep{da2022proving, burkholz2022convolutional, ferbach2023a}. 
    
    However, the theoretical foundation of the SLTH for \textit{transformers}, which form the basis of modern language models, remains unexplored---due to a transformer-specific component, \textit{an attention mechanism}.
    As shown in the bottom panel of \Cref{fig:outline} (right side), one of the distinctive structures in transformers is the inner product between two vectors called \textit{query} and \textit{key}, obtained as linear projections of given inputs.
    This structure fundamentally differs from the conventional components of non-transformer architectures for which the SLTH has been established (the top panel of \Cref{fig:outline}); thus, it remains a mystery whether transformers contain SLTs under existing theoretical insights.
    This gap motivates our key research question:
    \textit{does an attention mechanism---an essential component of transformers---contain an SLT?}

    In this work, we prove the existence of SLTs within attention mechanisms, extending the SLTH to transformers.
    More precisely, we prove a suitably pruned source attention mechanism with random weights can approximate any target attention mechanism with arbitrary weights:
    \begin{theorem}[informal]\label{thr:attn_approx_informal}
        Given inputs of length $T$, a suitably pruned randomly-initialized attention mechanism of the input dimension $d$ and hidden dimension ${n=O(d\log(d^{3/2}/\epsilon))}$ can approximate an arbitrary attention mechanism of the same input dimension with an approximation error $\epsilon$, with high probability.
    \end{theorem}
    Our key idea is to reinterpret the inner product between the query and key vectors in the attention mechanism as a (linear) neural network weighted by the query and key projection matrices.
    Then, we can view the source and target inner products as neural networks with different numbers of layers: 
    the source one has two layers with query and key projection matrices as its weights, while the target one has a single layer with a weight matrix obtained by merging these two projections.
    This reinterpretation makes it possible to apply a variant of the two-layers-for-one approximation, leading to the SLT existence within attention mechanisms (\Cref{thr:slt_existence_attn_mech}).
    Note that, as can be seen by comparing the top and bottom panels of \Cref{fig:outline}, our arguments do not require additional layers in the MHA for approximation, in contrast to the previous two-layers-for-one argument for fully-connected networks.
    By exploiting this theorem, we further establish the SLTH for transformers without normalization layers: a randomly-initialized transformer has an SLT that approximates an arbitrary transformer with similar structures
    (\Cref{thr:slt_existence_transformer}).
    
    We also empirically validate our theory and confirm its implications. Specifically, we show that 
    1) the approximation error between the source and target attentions (or, more generally, source and target transformers) decays exponentially as the hidden dimension increases; and 
    2) this approximation error does not diverge even when the input length $T$ increases. 
    Also, based on our theoretical arguments, we derive a new, practical weight initialization scheme, leading to better SLTs in our experiments.
        
    Our contributions are summarized as follows:
    \begin{itemize}
        \item
            We provide the first theoretical proof that SLTs exist within attention mechanisms and transformers by reinterpreting the inner product in attention mechanisms. 
        \item
            We then empirically validate our theory under conditions that are close to our theoretical assumptions.
            More precisely, we carefully designed a synthetic experiment to observe how the hidden dimension or input length affects the approximation error of SLTs.            
        \item 
            Furthermore, we demonstrate that our theory not only explains the empirical results, but also provides a new insight into a weight initialization for finding better SLTs in practical settings.

    \end{itemize}

    \paragraph{Notation:}
        In this paper, scalars, vectors, and matrices are denoted by lowercase, bold lowercase, and bold uppercase letters, respectively. 
        We use the norm of matrices and vectors $\|\cdot\|$ as the spectral norm unless otherwise specified by subscripts. 
        We denote the uniform distribution on $[a, b]$ by $U[a, b]$.
        "$\odot$" represents an element-wise multiplication (\ie, the Hadamard product).  
        The superscript $(i)$ denotes the layer index, and we write $\{x^{(i)}\}_{i=1}^{H}$ to denote the set of elements $x^{(i)}$ indexed by $i$ from 1 to $H$.

\section{Preliminaries}
    This section reviews the prior theoretical studies on the strong lottery ticket hypothesis (SLTH) and the formulation of multi-head attention (MHA) mechanisms.

    \subsection{Strong Lottery Ticket Hypothesis}
    The strong lottery ticket hypothesis (SLTH) conjectured that a randomly-initialized network inherently contains subnetworks (strong lottery tickets (SLTs)) that achieve high accuracy comparable to trained dense networks, without any weight updates~\citep{ramanujan2020s,malach2020proving}.
    The first theoretical result of the SLTH was given by~\citet{malach2020proving}.
    They proved the existence of SLTs in a fully-connected ReLU network.
    Subsequent studies relaxed the requirements for source networks to contain SLTs that approximate some target network~\citep{orseau2020logarithmic,pensia2020optimal,burkholz2022most}.
    In particular, \citet{pensia2020optimal} introduced a subset-sum approximation technique~\citep{lueker1998exponentially} into the SLTH context and concluded that the logarithmic overparameterization of the source network to a given target is approximately optimal:
    \begin{lemma}
    \label{lem:pensia_layer_approx}
        Given ${\bm{x} \in \mathbb{R}^{d_1}}$, ${\bm{W}\in \mathbb{R}^{d_2 \times d_1}}$, $\tilde{\bm{W}}_1 \in \mathbb{R}^{n \times d_1}$, and $\tilde{\bm{W}}_2 \in \mathbb{R}^{d_2 \times n}$, we define the target and pruned source fully-connected networks as
        \begin{align*}
            \tffn(\bm{x}) &:= \bm{W}\bm{x}, \\
            \sffn(\bm{x}) &:= (\tilde{\bm{W}}_2\odot \bm{M}_2)\relu((\tilde{\bm{W}}_1\odot \bm{M}_1)\bm{x}),
        \end{align*}
        where ${\bm{M}_1 \in \{0, 1\}^{n \times d_1}}$ and ${\bm{M}_2 \in \{0, 1\}^{d_2 \times n}}$ are binary pruning masks.
        Assume that ${\|\bm{W}\|\le 1}$, ${\|\bm{x}\| \le 1}$, and each entry of $\tilde{\bm{W}}_1$ and $\tilde{\bm{W}}_2$ is drawn i.i.d. from $U[-1, 1]$.
        Also, for $0<\epsilon<1$, suppose that the hidden dimension $n$ satisfies ${n\ge d_1 C \log\left(2d_1d_2/\epsilon\right)}$, where $C>0$ is some universal constant.
        Then, with probability at least $1 - \epsilon$, there exists a choice of binary pruning masks $\bm{M}_1$ and $\bm{M}_2$ such that
        \begin{align*}
            \|\tffn(\bm{x}) - \sffn(\bm{x})\| \le \epsilon.
        \end{align*}
    \end{lemma}
    This approach, which approximates a single weight matrix by pruning two randomly initialized matrices (the top panel of \Cref{fig:outline}), is called the two-layers-for-one approximation, and is now the theoretical foundation of the SLTH for more complex architectures and problems~\citep{da2022proving,burkholz2022convolutional,ferbach2023a,natale2024on,otsuka2025partially}.

    \begin{figure}[tb]
        \centering
        \includegraphics[width=0.95\columnwidth]{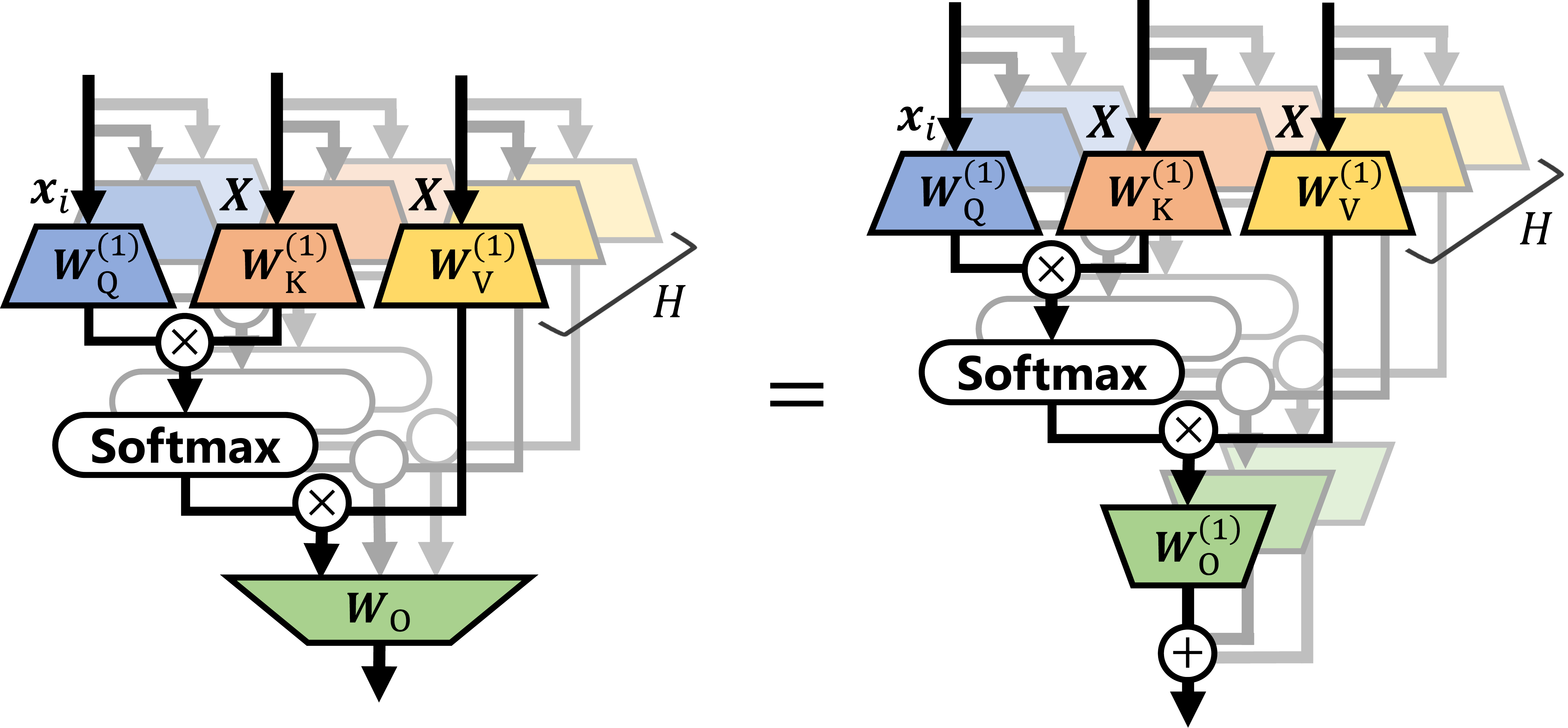}
        \caption{
            The structure of an MHA. 
            By partitioning the output projection, the final result can be interpreted as the sum of outputs from all heads.
            }
        \label{fig:multi_head_attention}
    \end{figure}

    \subsection{Multi-head Attention Mechanisms}

        Let ${\bm{X} = [\bm{x}_1, ..., \bm{x}_T]^\top \in \mathbb{R}^{T \times d_1}}$ be a sequence of $T$ input vector embeddings.
        For each embedding $\bm{x}_i$, we define a binary attention mask ${\bm{a}_i \in \{0,1\}^\top}$, where ${\bm{a}_{i,j} = 1}$ indicates that the $i$-th embedding attends to the $j$-th one.
        We assume that each embedding attends to at least one other (i.e., ${\|\bm{a}_i\|_1 \ge 1}$).
        Given such inputs, a multi-head attention (MHA) mechanism~\citep{vaswani2017attention} is defined as a function that computes their pair-wise relationships at each of the $H$ attention heads (the left panel of \Cref{fig:multi_head_attention}).
        For the $i$-th embedding $\bm{x}_i$, the MHA is of the following form:
        \begin{align*}
            \attn(\bm{x}_i; \bm{X}, \{&\bm{W}_{\mathrm{Q}}^{(j)}, \bm{W}_{\mathrm{K}}^{(j)}, \bm{W}_{\mathrm{V}}^{(j)} \}_{j=1}^{H}, \bm{W}_{\mathrm{O}}) \\
                :=& \left[\mathrm{head}_i^{(1)}, \dots, \mathrm{head}_i^{(H)}\right] \bm{W}_{\mathrm O} 
                \in \mathbb{R}^{1 \times d_2}, \\
                \mathrm{head}_i^{(j)} 
                :=& \softmax\left(
                    \frac{\bm{q}_i^{(j)} \bm{K}^{(j)\top}}{\sqrt{d_{\mathrm K}}}; \bm{a}_i
                \right) \bm{V}^{(j)}, \\
                \bm{q}_i^{(j)}
                := \bm{x}_i^\top \twquery^{(j)}, &\quad \bm{K}^{(j)}
                := \bm{X} \twkey^{(j)}, \quad \bm{V}^{(j)}
                := \bm{X} \twvalue^{(j)}, \\
                \softmax(\bm{x}_i; \bm{a}_i)_j 
                :=& \frac{a_{i, j} \exp(x_{i, j})}{\sum_{k=1}^T a_{i, k} \exp(x_{i, k})}.
            \end{align*}
        Here, we define ${\bm{W}_{\mathrm Q}^{(j)}, \bm{W}_{\mathrm K}^{(j)} \in \mathbb{R}^{d_1 \times d_{\mathrm K}}}, {\bm{W}_{\mathrm V}^{(j)} \in \mathbb{R}^{d_1 \times d_{\mathrm V}}}$, and ${\bm{W}_{\mathrm O} \in \mathbb{R}^{H d_{\mathrm V} \times d_2}}$ as single-layer projections for the \textit{query} $\bm{q}_i^{(j)}$, \textit{key} $\bm{K}^{(j)}$, \textit{value} $\bm{V}^{(j)}$, and output of the MHA, respectively.
        The softmax function with the attention mask is defined as $\softmax(\cdot)$.
        As shown in the right panel of \Cref{fig:multi_head_attention}, by partitioning the output weight matrix $\bm{W}_{\mathrm O}$ into 
        \[
            {\bm{W}_{\mathrm O} := [\bm{W}_{\mathrm O}^{(1)\top}, \dots, \bm{W}_{\mathrm O}^{(H)\top} ]^\top},\quad \bm{W}_{\mathrm O}^{(j)} \in \mathbb{R}^{d_{\mathrm V} \times d_2},
        \]    
        the form of $\attn(\cdot)$ can be represented as follows:
        \begin{align*}
        \attn(\bm{x}_i ; \bm{X}, \{\bm{W}_{\mathrm{Q}}^{(j)}, \bm{W}_{\mathrm{K}}^{(j)}, &\bm{W}_{\mathrm{V}}^{(j)} \}_{j=1}^{H}, \bm{W}_{\mathrm{O}}) \\
        &=\attn(\bm{x}_i; \bm{X}, \bm{W}_{\mathrm{Q:O}}^{(1:H)})\\
        &= \sum_{j=1}^H \mathrm{head}_i^{(j)} \bm{W}_{\mathrm O}^{(j)}.
        \end{align*}
        We denote the set of all weights as 
        \[
            {\bm{W}_{\mathrm{Q:O}}^{(1:H)} := \{\bm{W}_{\mathrm{Q}}^{(j)}, \bm{W}_{\mathrm{K}}^{(j)}, \bm{W}_{\mathrm{V}}^{(j)}, \bm{W}_{\mathrm{O}}^{(j)} \}_{j=1}^{H}}.
        \]
\section{Strong Lottery Ticket Hypothesis\\for Transformers}\label{sec:main_theorem}
    
    This section analyzes the existence of SLTs within multi-head attention (MHA) mechanisms and extends it to the transformer architecture without normalization layers.
    For a detailed proof, see \Cref{sec:proofs_app}.

    \subsection{Setups}\label{subsec:theorem_setup}            
                 We consider two MHAs: a target MHA $\tattn(\cdot)$ with arbitrary (tuned) weights, and a pruned source MHA $\sattn(\cdot)$ with randomly-initialized weights, denoted as follows:
                \begin{align}\label{eq:def_sattn}
                    \tattn(\bm{x}_i) 
                    &= 
                    \attn(\bm{x}_i; \bm{X}, \twset), \\
                    \label{eq:def_tattn}
                    \sattn(\bm{x}_i) 
                    &= 
                    \attn(\bm{x}_i; \bm{X}, \swset).
                \end{align}
                Here, similarly to the weight set $\twset$, we define the set of pruned random weights as
                \begin{align*}
                    \swset 
                    &:= 
                    \{\tilde{\bm{W}}_{\mathrm Q}^{(j)} \odot \bm{M}_{\mathrm Q}^{(j)}, \tilde{\bm{W}}_{\mathrm K}^{(j)} \odot \bm{M}_{\mathrm K}^{(j)}, \\
                    &\quad\quad\quad \tilde{\bm{W}}_{\mathrm V}^{(j)} \odot \bm{M}_{\mathrm V}^{(j)}, \tilde{\bm{W}}_{\mathrm O}^{(j)} \odot \bm{M}_{\mathrm O}^{(j)} \}_{j=1}^H,
                \end{align*}
                where ${\swquery^{(j)}, \swkey^{(j)} \in \mathbb{R}^{d_1 \times n_{\mathrm{K}}}}$, ${\swvalue^{(j)} \in \mathbb{R}^{d_1 \times n_{\mathrm{V}}}},$ and ${\swout^{(j)} \in \mathbb{R}^{n_{\mathrm{V}} \times d_2}}$ are the randomly-weighted query, key, value, and output projections of the $j$-th head in $\sattn(\cdot)$, respectively. 
                Also, $\smquery^{(j)}$, $\smkey^{(j)}$, $\smvalue^{(j)}$, and ${\smout^{(j)}}$ are their corresponding binary pruning masks.
                Note that the target and source MHAs have different key and value hidden dimensions: $d_{\mathrm{K}}$ and $ d_{\mathrm{V}}$ for the target, and $n_{\mathrm{K}}$ and $n_{\mathrm{V}}$ for the source.
                We assume that $\alpha \ge \max(\sqrt{d_1}, \sqrt{d_2})$ for the inputs, and $\|\twquery^{(j)}\|, \|\twkey^{(j)}\|, \|\twvalue^{(j)}\|, \|\twout^{(j)}\|\le 1$ for the $j$-th head of the target MHA.
                The source MHA is initialized such that each entry of $\swquery$ and $\swkey$ is drawn i.i.d. from $U[-n_{\mathrm{K}}^{1/4}, n_{\mathrm{K}}^{1/4}]$, and each entry of $\swvalue$ and $\swout$ is drawn i.i.d. from $U[-1, 1]$.

        \subsection{The Existence of SLTs Within an MHA}        
            Now, we prove the following SLT existence theorem:
            \begin{theorem}\label{thr:slt_existence_attn_mech}
                Let $\tattn(\cdot)$ and $\sattn(\cdot)$ be as defined in \Cref{eq:def_sattn,eq:def_tattn}. Then, with probability at least $1 - \epsilon$, there exists a choice of binary pruning masks $\bm{M}_{\mathrm Q}^{(j)}, \bm{M}_{\mathrm K}^{(j)}, \bm{M}_{\mathrm V}^{(j)}, \bm{M}_{\mathrm O}^{(j)}$ that satisfy
                \begin{align*}
                    \max_{i \in [T]} \|\sattn(\bm{x}_i) - \tattn(\bm{x}_i)\| \le \epsilon,
                \end{align*}
                if the source hidden dimensions satisfy
                \begin{align*}
                    n_{\mathrm{K}} &\ge d_1 C \log\left(\frac{8H\alpha^3 d_1^{3/2}}{\epsilon}\right), \\
                    n_{\mathrm{V}} &\ge d_1 C \log\left(\frac{2H\alpha d_1 \sqrt{d_2}}{\epsilon}\right),
                \end{align*}
                for some universal constant $C > 0$.
            \end{theorem}

\begin{figure}[tb]
    \centering
    \includegraphics[width=0.85\linewidth]{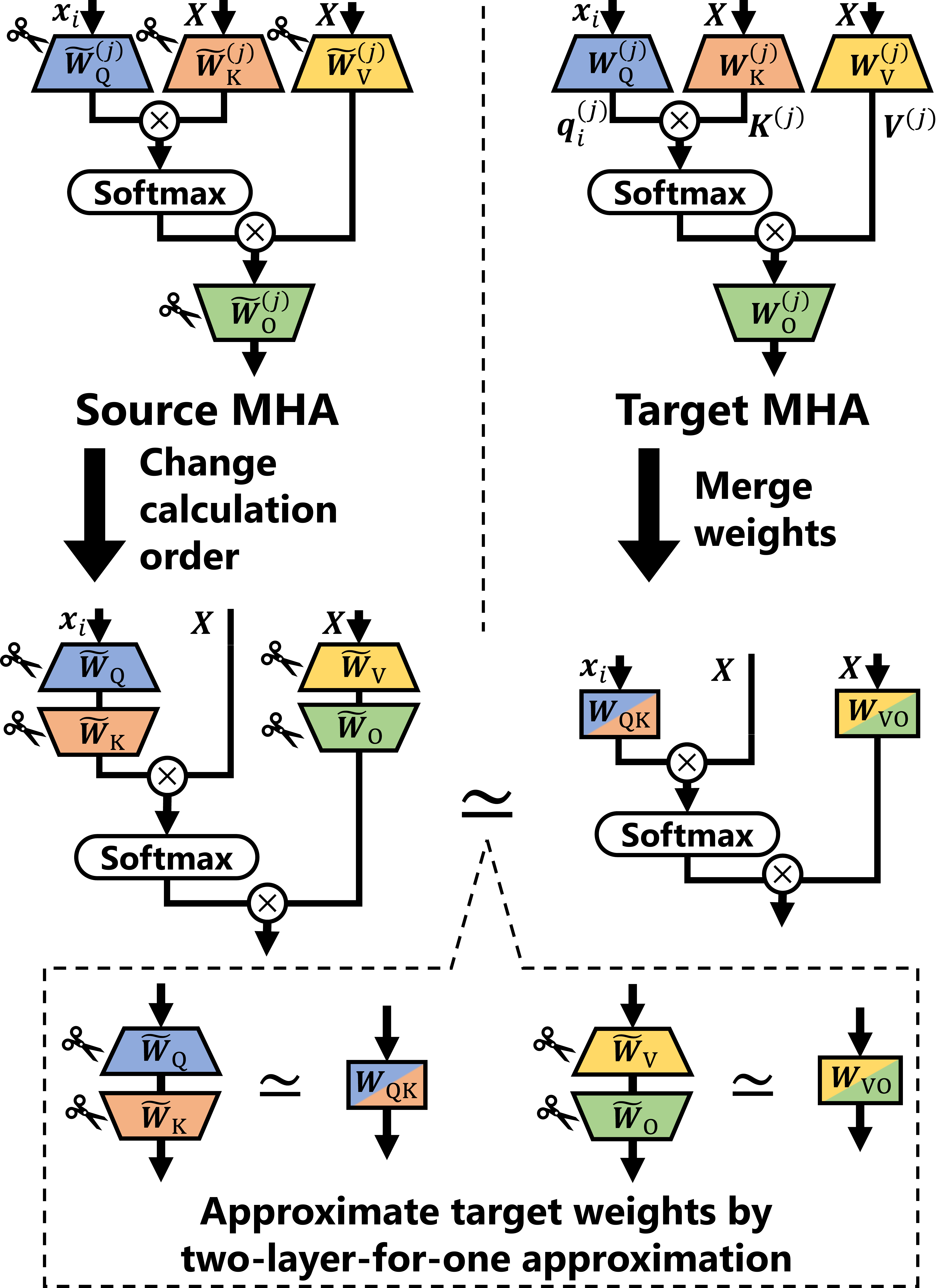}
    \caption{
        The diagram of our proof.
        By merging the target projections and changing the calculation order of the source MHA,
        we can apply a variant of the two-layers-for-one approximation technique and approximate the target MHA while keeping the original source and target structures.
        }
    \label{fig:proof_step}
\end{figure}

            \Cref{fig:proof_step} shows an overview of our proof.
            To prove \Cref{thr:slt_existence_attn_mech}, we begin by focusing on the part before the softmax, the target and source inner products for the $j$-th head:
            \begin{align}
                \label{eq:part_before_t_softmax}
                \frac{1}{\sqrt{d_\textrm{K}}}\bm{q}^{(j)}\bm{K}^{(j)\top} = \frac{1}{\sqrt{d_\textrm{K}}}(\bm{x}_i^\top \twquery^{(j)})(\bm{X} \twkey^{(j)})^\top, \\
                \label{eq:part_before_s_softmax}
                \frac{1}{\sqrt{n_\textrm{K}}}(\bm{x}_i^\top (\swquery^{(j)} \odot \smquery^{(j)}))(\bm{X} (\swkey^{(j)} \odot \smkey^{(j)}))^\top.
            \end{align}
            Since the only difference lies in the projection matrices, we consider the problem of pruning the source projections $\swquery^{(j)}$ and $\swkey^{(j)}$ to approximate the target projections $\twquery^{(j)}$ and $\twkey^{(j)}$.
            A naive idea might be to approximate each target projection independently.
            In this case, a single source random matrix must approximate each target matrix.
            However, pruning a single random matrix cannot generally approximate arbitrary ones;
            thus, this approach is infeasible.
            To overcome this limitation, we revisit the structure of the target inner product.
            By closely examining the formulation of the target inner product (\Cref{eq:part_before_t_softmax}), we observe that the query and (transposed) key projections appear adjacently and can be merged into a single joint projection (the right panel of \Cref{fig:proof_step}):
            \begin{align}\label{eq:W_qk_W_vo} \nonumber
                \frac{1}{\sqrt{d_\textrm{K}}}(\bm{x}_i^\top \twquery^{(j)})(\bm{X} \twkey^{(j)})^\top 
                =& \ 
                \bm{x}_i^\top \bm{W}_{\mathrm{QK}}^{(j)}  \bm{X}^\top, \\
                \bm{W}_{\mathrm{QK}}^{(j)} :=& \frac{1}{\sqrt{d_k}}\bm{W}_{\mathrm{Q}}^{(j)}(\bm{W}_{\mathrm K}^{(j)})^\top.
            \end{align}
            This reformulation enables us to reinterpret the original problem---not as approximating two target matrices---but as approximating a single merged projection matrix.

            We now approximate this merged matrix $\bm{W}_{\mathrm{QK}}^{(j)}$ by pruning the two source projections.
            On the source side (\Cref{eq:part_before_s_softmax}) as well, the query and key projections are adjacent.
            Thus, the source inner product can be viewed as a computation that first calculates the query and key projections
            (the left panel of \Cref{fig:proof_step}):
            \begin{align*}
                \frac{1}{\sqrt{n_\textrm{K}}}(\bm{x}_i^\top (&\swquery^{(j)} \odot \smquery^{(j)}))(\bm{X} (\swkey^{(j)} \odot \smkey^{(j)}))^\top \\
                =& \ 
                \bm{x}_i^\top \left((\swquery'^{(j)} \odot \smquery^{(j)})(\swkey'^{(j)} \odot \smkey^{(j)})^\top\right) \bm{X}^\top, \\
                \swquery'^{(j)}&
                :=
                \frac{1}{n_\textrm{K}^{1/4}}\swquery^{(j)}, \quad
                \swkey'^{(j)}
                :=
                \frac{1}{n_\textrm{K}^{1/4}}\swkey^{(j)}, 
            \end{align*}
            where each entry of $\swquery'^{(j)}$ and $\swkey'^{(j)}$ is drawn i.i.d. from $U[-1, 1]$ as per our assumption.
            Therefore, the task reduces to selecting masks $\smquery^{(j)}$ and $\smkey^{(j)}$ such that the source matrix product ${(\swquery'^{(j)} \odot \smquery^{(j)})(\swkey'^{(j)} \odot \smkey^{(j)})^\top}$ closely approximates the target $\bm{W}_{\mathrm{QK}}^{(j)}$.
            This allows us to draw an analogy to the conventional theoretical results of the SLTH, particularly the two-layers-for-one approximation (\Cref{lem:pensia_layer_approx}).
            We therefore establish and apply a variant of \Cref{lem:pensia_layer_approx}, which guarantees the existence of binary pruning masks that achieve such an approximation (the bottom panel of \Cref{fig:proof_step}):
            \begin{lemma}\label{lem:weight_approximation}
                Let ${\bm{W} \in \mathbb{R}^{d_2 \times d_1}}$ be a target matrix with ${\|\bm{W}\|\le 1}$, and $\tilde{\bm{W}}_1 \in \mathbb{R}^{n \times d_1}$ and $\tilde{\bm{W}}_2 \in \mathbb{R}^{d_2 \times n}$ be source matrices whose entries are drawn i.i.d. from $U[-1, 1]$.
                Suppose that ${n \ge d_1 C \log(d_1 d_2/\epsilon)}$ for some universal constant $C > 0$.
                Then, with probability at least $1 - \epsilon$, there exists a choice of binary pruning masks $\bm{M}_1$ and $\bm{M}_2$ such that
                \[
                    \left\| \bm{W} - (\tilde{\bm{W}}_2 \odot \bm{M}_2)(\tilde{\bm{W}}_1 \odot \bm{M}_1) \right\|_{\max} \le \frac{\epsilon}{d_1 d_2}.
                \]
            \end{lemma}

            We now turn to the components after the softmax function: the value and output projections.
            Similar to the query and key case, the value and output projections appear adjacently and can also be merged into a single composite transformation. 
            Thus, we aim to approximate the target merged matrix $\bm{W}_{\mathrm{VO}}^{(j)} := \bm{W}_{\mathrm{V}}^{(j)}\bm{W}_{\mathrm{O}}^{(j)}$. 
            This approximation follows the same principle as before: we leverage the matrix product on the source side $(\swvalue^{(j)} \odot \smvalue^{(j)})(\swout^{(j)} \odot \smout^{(j)})$ to approximate the merged matrix $\bm{W}_{\mathrm{VO}}^{(j)}$.
            \Cref{lem:weight_approximation} ensures that, with high probability, this approximation is successful via appropriately chosen binary pruning masks $\smvalue^{(j)}$ and $\smout^{(j)}$.

            Assuming that all weights in the target MHA are approximated by the above procedure, we next analyze the error of the entire attention mechanism by 
            investigating the behavior of the softmax.
            As a natural idea, one might consider exploiting the 1-Lipschitz continuity of the softmax~\citep{gao2017properties}, which enables internal errors to propagate linearly to the output.
            However, since the MHA subsequently multiplies the softmax output and the input matrix $\bm{X}$, applying Lipschitz continuity results in a loose upper bound of the error between MHAs: as $\|\bm{X}\|$ can grow with $T$ in the worst case, the bound depends on the input length $T$.

            In contrast to this general approach, we provide a more precise analysis.
            In our setting, thanks to the accurate weight approximation technique mentioned earlier, the internal error of softmax is guaranteed to be finite and small.
            Leveraging this property, we analyze the softmax output and $\bm{X}$ simultaneously to obtain a $T$-independent bound as follows:
            \begin{lemma}\label{lem:softmax_diff_bound}
                    Let ${\bm{\epsilon} \in \mathbb{R}^{d_1}}$ be an error vector with ${\|\bm{\epsilon}\|_{\max} \leq \epsilon_{\max}}$ for some ${0\le\epsilon_{\max} \le 1/2}$. 
                    Then,
                    \[
                        \max_{i\in[T]}\left\| \softmax(\bm{x}_i; \bm{a}_i) \bm{X} - \softmax(\bm{x}_i + \bm{\epsilon}; \bm{a}_i) \bm{X} \right\| 
                        \leq 
                        4\sqrt{d_1} \alpha \epsilon_{\max}.
                    \]
            \end{lemma}
            Since this lemma provides a bound independent of $\bm{a}_i$, our theory holds for models with arbitrary attention masks, including encoder~\citep{devlin2019bert} and decoder models~\citep{radford2019language}.
            By applying these above analyses to each attention head, we complete the proof of \Cref{thr:slt_existence_attn_mech}.
            For the full proof,
            see \Cref{subsec:SLT_existence_within_attention_mechanisms_app}.
            We also empirically validate two main theoretical findings in \Cref{subsec:main_theorem_verification}:
            the accurate approximation of the target MHA becomes feasible with larger source hidden dimensions, and the the approximation error remains independent of the input length $T$.

            \paragraph{Proof Sketch of \Cref{thr:slt_existence_attn_mech}:}
            First, for each attention head, we reformulate the problem by merging the four original target projection matrices into two merged matrices: one combining the query and key projections, and the other the value and output projections.
            Applying \Cref{lem:weight_approximation} to these merged matrices enables us to prune each source head to produce an inner product that closely approximates the target one.
            Next, using \Cref{lem:softmax_diff_bound}, we bound how errors in approximating the query and key matrices propagate through the softmax operation. 
            \Cref{lem:softmax_diff_bound} ensures that the approximation error of the softmax depends on the approximation accuracy of the query-key projections and does not scale with the input length $T$.
            Thus, provided the source hidden dimensions $n_{\mathrm{K}}$ and $n_{\mathrm{V}}$ are sufficiently large, there exists a choice of binary masks for the source MHA which approximate the target MHA within an error $\epsilon$.
            Also, by suitably setting lower bounds on $n_{\mathrm{K}}$ and $n_{\mathrm{V}}$, a union bound guarantees that the approximation succeeds across all heads with probability at least $1 - \epsilon$.

            \subsection{The Existence of SLTs Within a Transformer}\label{subsec:slt_existence_transformer}
            By leveraging our main theorem, we now extend the SLTH to transformers.
            We consider a transformer without the normalization layers for the original definition~\citep{vaswani2017attention}.
            The target transformer of $B$ blocks are of the following form:
            \begin{align*}
                \ttrans(\bm{x}_i) &:= \tblk^{(B)}(\tblk^{(B-1)}\dots \tblk^{(1)}(\bm{x}_i)), \\
                \tblk^{(b)}(\bm{x}_i^{(b)}) &:= \tffn^{(b)}(\tattn(\bm{x}^{(b)}_i)^\top + \bm{x}_i^{(b)}) \\
                &\qquad\qquad + \tattn(\bm{x}_i^{(b)})^\top + \bm{x}_i^{(b)},
            \end{align*}
            where $\tblk^{(b)}$ is a $b$-th target block, and $\bm{x}_i^{(b)}\in\mathbb{R}^d$ is the $i$-th input embedding of the $b$-th target block. 
            We employ single-layer projection $\tffn^{(b)} (\cdot)$ for the fully-connected network of each target block.
            Similarly, we define the pruned source transformer as follows:
            \begin{align*}
                \strans(\bm{x}_i) &:= \sblk^{(B)}(\sblk^{(B-1)}\dots \sblk^{(1)}(\bm{x}_i)), \\
                \sblk^{(b)}(\bm{x}_i'^{(b)}) &:= \sffn^{(b)}(\sattn(\bm{x}_i'^{(b)})^\top + \bm{x}_i'^{(b)}) \\
                &\qquad\qquad + \sattn(\bm{x}_i'^{(b)})^\top + \bm{x}_i'^{(b)},
            \end{align*}
            where $\sblk^{(b)}(\cdot)$ is a $b$-th source block and $\bm{x}_i'^{(b)}$ is the $i$-th input embedding of the $b$-th source block.
            We set the hidden dimension of $\sffn^{(b)}(\cdot)$ as $n_{\textrm{FC}}^{(b)}$
            and assume the hidden dimensions of the MHA in the $b$-th source block as a same value $n_{\textrm{MHA}}^{(b)}$ for simplicity. 
            Then, we prove the following theorem:
            \begin{theorem}\label{thr:slt_existence_transformer}
                Assume $B\ge 2$. Then, with probability at least $1 - \epsilon$ for $0 < \epsilon < 1$, there exists a choice of binary pruning masks that satisfies
                \begin{align*}
                    \|\strans(\bm{x}_i) - \ttrans(\bm{x}_i)\| \le \epsilon,
                \end{align*}
                if the hidden dimensions of $b$-th source MHA and fully-connected network satisfy
                \begin{align*}
                    n_{\mathrm{MHA}}^{(b)}
                    &\ge
                    d_1 C\log\left( \frac{c_1^{f_1(b, B)} H^{f_2(b, B)} d_1^{f_3(b, B)}}{\epsilon} \right), \\
                    n_{\mathrm{FC}}^{(b)}
                    &\ge
                    d_1 C\log\left( \frac{c_2^{g_1(b, B)} H^{g_2(b, B)} d_1^{g_3(b, B)}}{\epsilon} \right),
                \end{align*}
                for universal constants $C>0$ and ${c_1, c_2>0}$ including $\alpha$.
                Here, $f_1, f_2, f_3, g_1, g_2, g_3$ are quadratic forms of $b$ and $B$.
            \end{theorem}
            \paragraph{Proof Sketch of \Cref{thr:slt_existence_transformer}:}          
                From the existing work (\Cref{lem:pensia_layer_approx}) and \Cref{thr:slt_existence_attn_mech}, we already know that an MHA and FFN contain SLTs with high probability if each module has a large hidden dimension; 
                thus, by determining the lower bound of the hidden dimension of each module based on the error propagation from the input to output, we can prove that there exists an SLT, which approximates the output of an target transformer to an error of $\epsilon$, within a randomly initialized transformer.
                By the union bound, the probability that all approximations hold simultaneously is at least $1- \epsilon$.
            
            For simplicity, this theorem uses target and source fully-connected networks as a single-layer and two-layer ReLU networks $\tffn$ and $\sffn$ in \Cref{lem:pensia_layer_approx}.
            It can be generalized to an $L$-layer target fully-connected network by applying the multi-layer approximation by \citet{pensia2020optimal}.
            We show that theorem and its proof in \Cref{subsec:theorem_transformer}.

\section{Experimental Results}
    This section empirically validates our SLTH theorems.

    \subsection{Experimental Settings}
        To empirically validate the approximation guarantees established by our SLTH theorems, we evaluate the approximation error on a synthetic dataset for angular velocity estimation. 
        The input consists of a sequence of two-dimensional vectors arranged on the unit circle with a fixed angular velocity. 
        A regression token is used to estimate this velocity, and the source model uses the same regression token as the target model to ensure input consistency.
        The source and target models are both implemented as either single-head attention mechanisms or single-head transformers as defined in \Cref{subsec:slt_existence_transformer}.
        Both models are initialized according to our theoretical setup: 
        the entries of the query and key projection weights are drawn i.i.d. from $U[-n_{\textrm{K}}^{1/4}, n_{\textrm{K}}^{1/4}]$, and those of the value and output projection weights from $U[-1, 1]$.
        To identify SLTs that approximate the target network, we implement the weight approximation technique described in \Cref{lem:weight_approximation}, which is based on the subset-sum approximation of \citet{pensia2020optimal}. 
        The target MHA is approximated using 100 randomly initialized source MHAs, and we report the mean and standard deviation of the approximation error. 
        
        We also investigate whether our theoretical insights generalize to practical settings. 
        In this setting, we search for SLTs by the \texttt{edge-popup} algorithm~\citep{ramanujan2020s}, which finds accurate subnetworks by backpropagation, instead of learning weights. 
        We train models from the GPT-2 family (mini\footnote{A 4-layer GPT-2. For details, see the following repository: https://huggingface.co/erwanf/gpt2-mini~\citep{wolf-etal-2020-transformers}}, small, and medium)~\citep{radford2019language} on the WikiText-103 dataset~\citep{merity2017pointer}. 
        The weights of these models are initialized based on the GPT-2 initialization scheme.
        For each model, we repeat training three times with different random seeds and report the mean and standard deviation of the final performance.
        See \Cref{app:experimental-details} for further details on experimental settings.

    \subsection{Verification of Main Theorems}\label{subsec:main_theorem_verification}
        We empirically verify our theoretical results by pruning a source network to approximate the target network.

        \begin{figure}[tb]
            \centering
            \includegraphics[width=0.9\columnwidth]{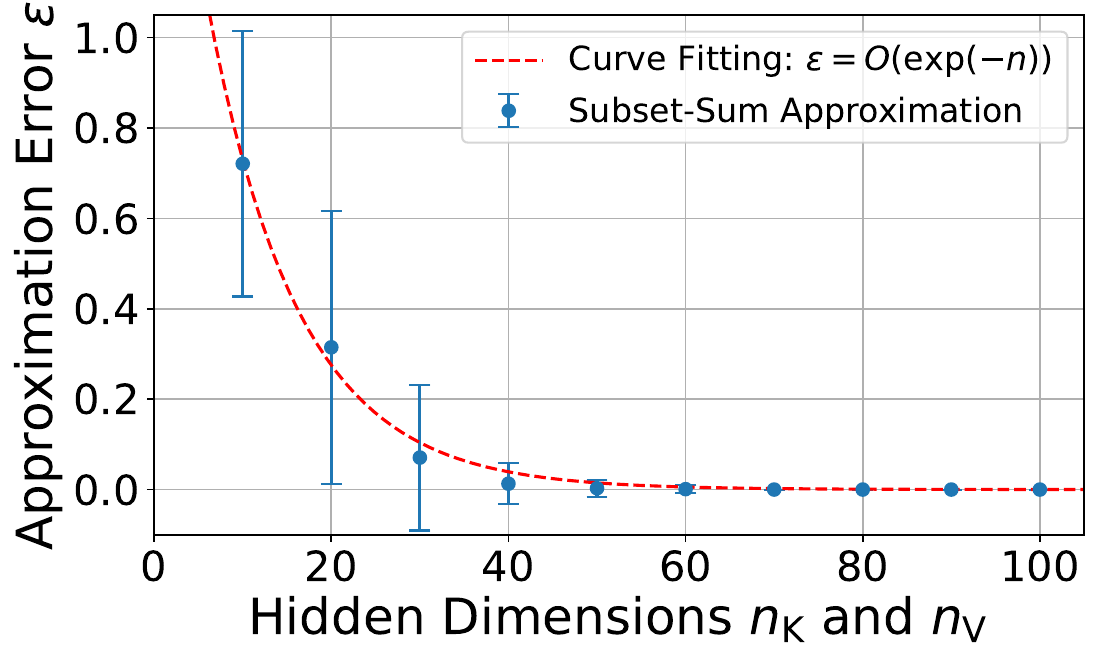}
            \caption{
                The approximation error $\epsilon$ of SLTs within a source MHA for the hidden dimensions ${n_{\mathrm{K}} = n_{\mathrm{V}}}$.
                This result shows that the error $\epsilon$ satisfies $\epsilon = O(\exp(-n))$, consistently with \Cref{thr:slt_existence_attn_mech}.
                }
            \label{fig:eps_dim_comperison}
        \end{figure}

        \paragraph{Varying the Hidden Dimensions:}
            We validate \Cref{thr:slt_existence_attn_mech} by showing that increasing the hidden dimensions leads to an exponential decrease in approximation error.
            When we fit the empirical results to the function ${\epsilon = \gamma \exp(-\delta n_{\mathrm{K}})}$, we obtain $\epsilon = 0.8 \exp(-0.06 n_{\mathrm{K}})$, which closely matches the observations.
            This finding supports our theoretical claim: given a target MHA, each source hidden dimension requires 
            $O(\log(1/\epsilon))$ for the existence of SLTs.

        \begin{figure}[tb]
            \centering
            \includegraphics[width=0.9\columnwidth]{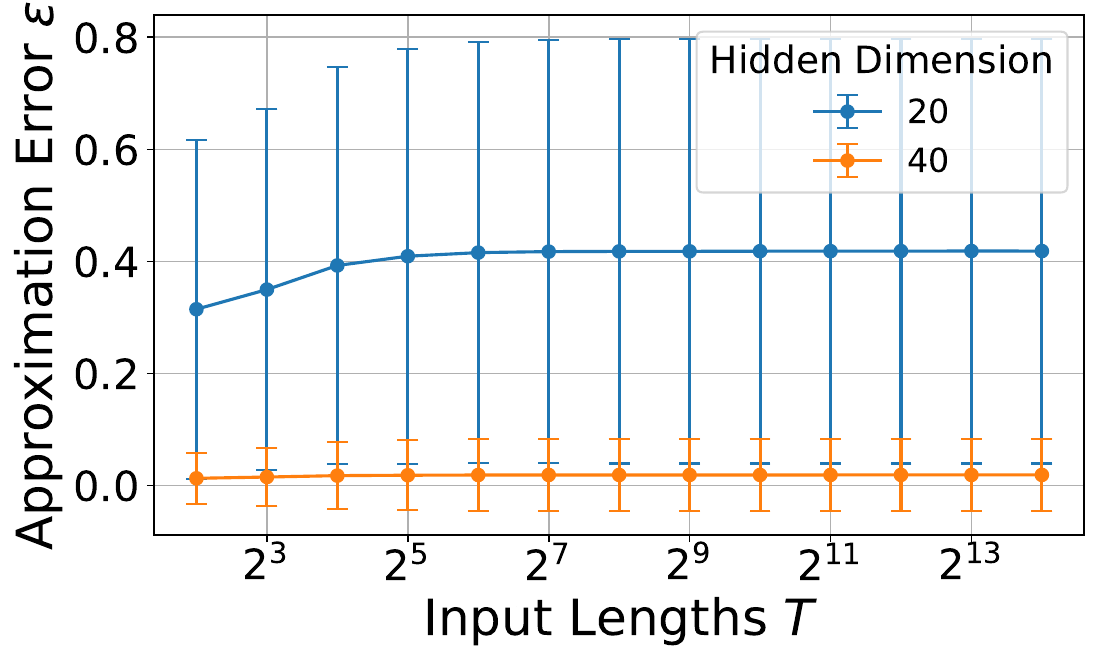}
            \caption{
                The approximation error $\epsilon$ of SLTs within an MHA for the sequence length $T$.
                This result suggests that the error $\epsilon$ does not diverge as $T$ increases, as implied by \Cref{thr:slt_existence_attn_mech}.
                }
            \label{fig:eps_seq}
        \end{figure}

        \paragraph{Varying the Sequence Length:}
            \Cref{thr:slt_existence_attn_mech} also implies that the existence of SLTs in MHAs is independent of the input length $T$.
            In other words, with sufficiently large hidden dimensions, the approximation error has an upper bound that does not depend on $T$.
            \Cref{fig:eps_seq} empirically supports this argument: even as $T$ increases, the error remains bounded, and the bound decreases with larger hidden dimensions.

        \begin{figure}[tb]
            \centering
            \includegraphics[width=0.9\columnwidth]{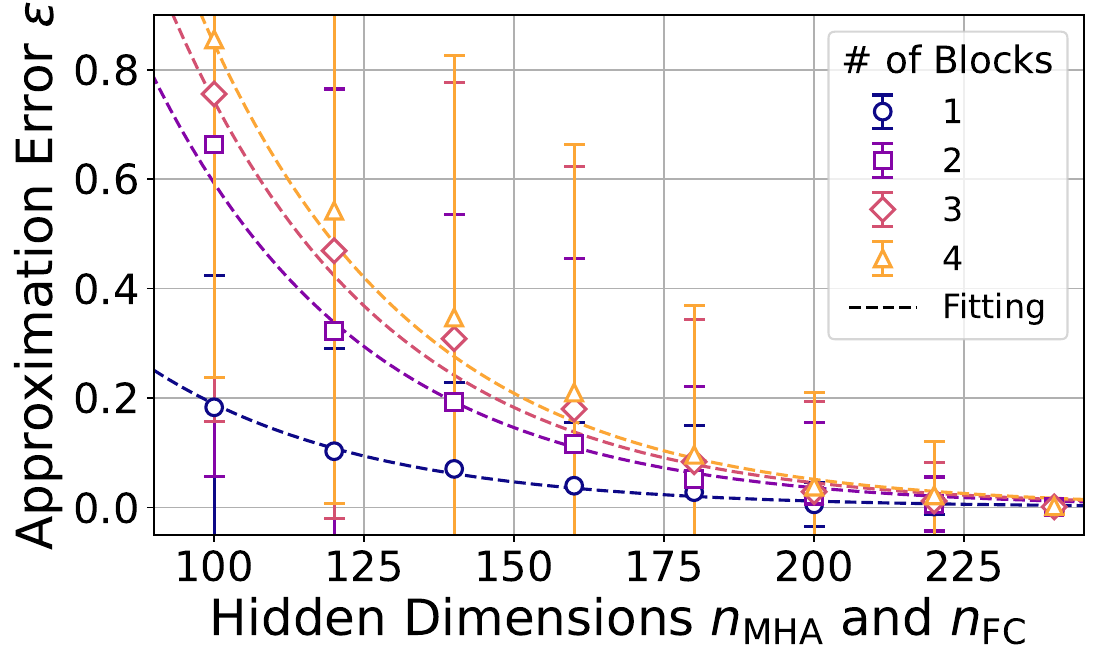}
            \caption{
                The approximation error $\epsilon$ of SLTs within a randomly initialized transformer for the source hidden dimensions ${n_{\mathrm{MHA}} = n_{\mathrm{FC}}}$.
                This result suggests that error accumulates as the number of blocks increases, while each error holds $\epsilon = O(\exp(-n_{\textrm{MHA}}))$, consistently with \Cref{thr:slt_existence_attn_mech}.
                }
            \label{fig:eps_block}
        \end{figure}

        \paragraph{Varying the Number of Blocks:}
            To validate \Cref{thr:slt_existence_transformer}, we analyze how the approximation error behaves across different numbers of transformer blocks. 
            We set $n_{\textrm{MHA}} = n_{\textrm{FFN}}$ and use an untrained target model to be close to our theoretical assumptions.
            As in the MHA experiment, we fit an exponential decay $\epsilon = \gamma \exp(-\delta n_{\mathrm{K}})$ to the error of each block, using the same decay rate $\delta$ obtained from the first block, as predicted by \Cref{thr:slt_existence_transformer}.
            \Cref{fig:eps_block} shows that, consistent with our theoretical implication, the approximation error decreases rapidly with increasing hidden dimensions for all numbers of blocks.
            Despite fitting only the coefficient $\gamma$ per block, the shared $\delta$ provides curves that closely match the empirical results, supporting our theoretical claim that only the scale factor varies across blocks.

    \begin{figure}[tb]
        \centering
        \includegraphics[width=0.9\columnwidth]{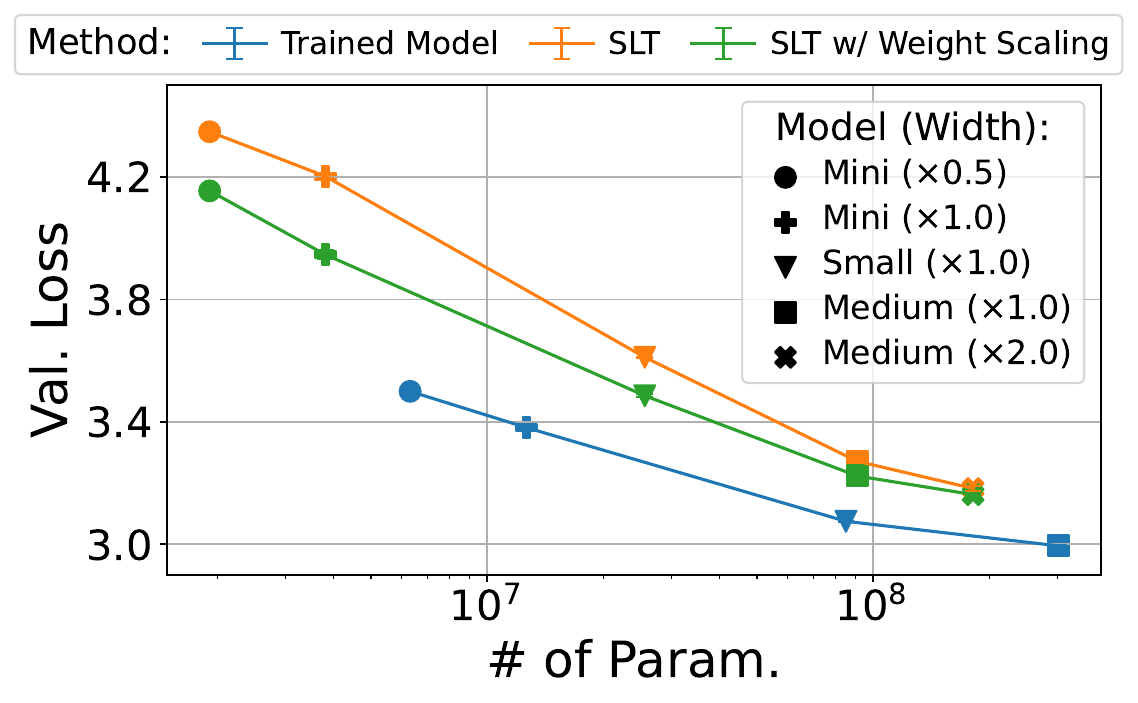}
        \caption{
            Loss comparison between SLTs with and without query and key weight scaling.
            By introducing a scale based on our theoretical assumptions, we can obtain better SLTs.
            }
        \label{fig:gpt_loss_param_comperison}
    \end{figure}

    \subsection{Behavior of SLTs in Practical Settings}
\begin{figure}[tb]
        \centering
        \includegraphics[width=0.92\columnwidth]{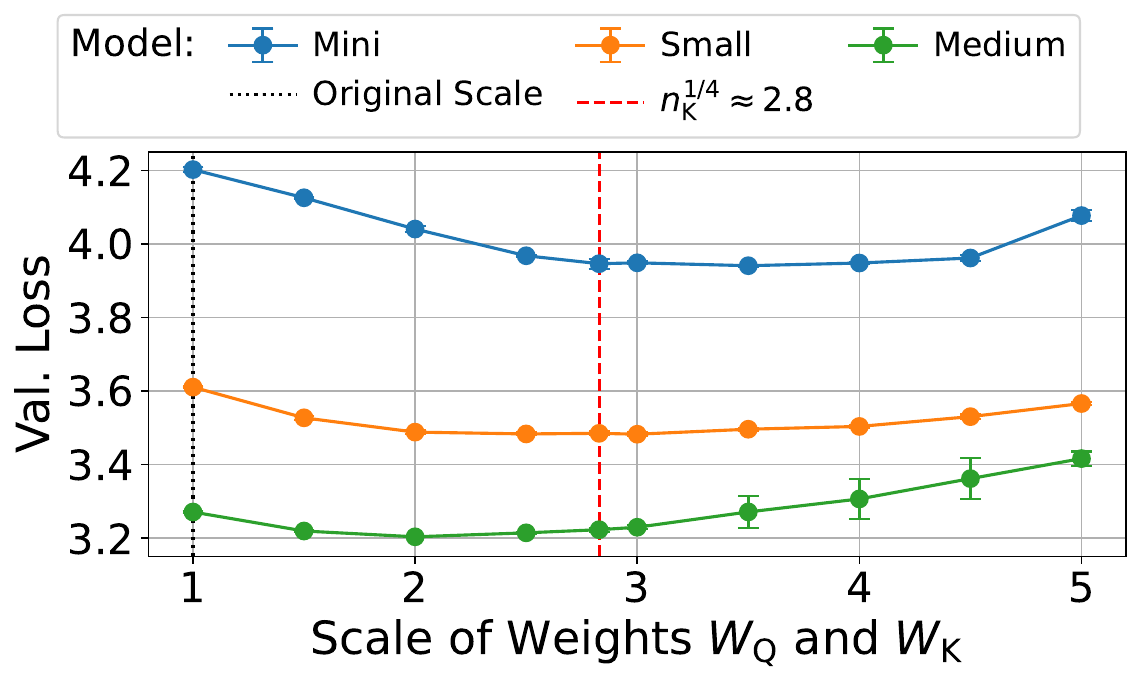}
        \caption{
            Loss comparison with respect to the weight scaling factor applied to query and key weights. 
            Interestingly, in all models, the loss reaches its minimum near the weight scaling of our theoretical assumptions.
            }
        \label{fig:gpt_scale_comperison}
    \end{figure}

        In the theoretical analysis, we employ a non-conventional initialization strategy: 
        the query and key projection weights are initialized from $U[-n_{\textrm{K}}^{1/4}, n_{\textrm{K}}^{1/4}]$, scaled by a factor of $n_{\textrm{K}}^{1/4}$ compared to the value and output weights, which are initialized from $U[-1, 1]$.
        This weight scaling was introduced to facilitate the application of the weight approximation lemma in our analysis, and played an important role in establishing our theory. 
        Its theoretical contribution motivates the following question: 
        \textit{does this scaled initialization strategy also benefit SLTs in realistic scenarios?}
        We empirically evaluate SLTs using the GPT-2 architectures and the WikiText-103 dataset.
        \Cref{fig:gpt_loss_param_comperison} compares the validation loss of SLTs with and without scaling the query and key weights by $n_{\textrm{K}}^{1/4} \simeq 2.8$, with respect to the number of nonzero parameters. 
        We observe that SLTs with the weight scaling tend to exhibit lower loss, approaching the performance of trained models. 
        Interestingly, this specific scaling factor $n_{\textrm{K}}^{1/4}$ is nearly optimal for finding better SLTs: 
        as shown in \Cref{fig:gpt_scale_comperison}, increasing the scale from 1 gradually decreases the loss up to a certain point, but further increasing it beyond $n_{\textrm{K}}^{1/4}$ results in increased loss. 
        In all models, the lowest loss is consistently achieved around this scaling factor $n_{\textrm{K}}^{1/4}$.
        These findings suggest that our initialization strategy actually helps to ensure the existence of better SLTs within the practical transformer models.

\section{Related Work}
    \paragraph{Strong Lottery Tickets:}
        \citet{zhou2019deconstructing} and \citet{ramanujan2020s} empirically found the subnetworks that achieve high accuracy without any weight training.
        The existence of such high-performing subnetworks has been called the strong lottery ticket hypothesis (SLTH), and its theoretical proof was firstly provided in fully-connected ReLU networks~\citep{malach2020proving,orseau2020logarithmic,pensia2020optimal,burkholz2022most}.

        Based on these pioneering studies, the SLTH has been extended in three main directions.
        The first direction involves introducing additional flexibility for relaxing the overparameterization of the source network~\citep{chijiwa2021pruning,xiong2023strong}.
        The second direction, in contrast, imposes additional constraints on the source network~\citep{gadhikar2023random,otsuka2025partially,natale2024on}. 
        The third direction extends the SLTH to various architectures~\citep{diffenderfer2021multiprize,burkholz2022most,fischer2021towards,da2022proving,da2023polynomially,burkholz2022convolutional,ferbach2023a}.
        Our work contributes to this third direction by proving the SLTH for attention mechanisms and transformers.

    \paragraph{Randomly Weighted Transformers:}
        Several studies have empirically investigated the capabilities of randomly weighted transformers.
        \citet{shen-etal-2021-reservoir} demonstrated that a transformer with a few randomly weighted layers achieves accuracy comparable to fully trained models on translation and language understanding tasks.
        \citet{zhong2024algorithmic} found that randomly weighted transformers can solve toy tasks with high accuracy as the hidden dimension increases.
        Some studies empirically showed the existence of SLTs within randomly weighted transformers~\citep{shen-etal-2021-whats,ito2025uncovering}.
        Our analysis provides theoretical support for these empirical results about the SLT existence. 
        Furthermore, it provides a theoretical explanation for the improved performance of randomly weighted transformers as the hidden dimension increases, particularly when the pruning is used for optimization.
\section{Conclusion}
    This work investigated the existence of SLTs within a multi-head attention (MHA) mechanism.
    We extended the existing theory of the SLTH to MHAs and proved that, if the source MHA has logarithmically large hidden dimensions, it contains an SLT that approximates an arbitrary MHA with high probability.
    Our proof revealed that, for the SLTH in MHAs, additional layers are not required for approximation, in contrast to the existing theories that rely on approximating a single-layer structure by a two-layer one.
    Furthermore, by exploiting our findings, we established the theory of the SLTH for transformers without normalization layers.
    We empirically validated our theory and confirmed that the results are consistent with the theoretical implications.
    Interestingly, our theoretical implication, which provides an appropriate weight scale for initializing query and key projection weights, contributed to improving the performance of SLTs in practical settings.
    Our results not only extend SLTH to transformers, but also indicate a new research direction in the SLTH for practical transformer models.
    We hope these findings will lead to a fundamental understanding of overparameterized models.

\section*{Acknowledgments}
This work was supported in part by JSPS KAKENHI Grant Number JP23H05489, JP25K03092, JP25KJ1236, and JST-ALCA-Next Japan Grant \# JPMJAN24F3.

\bibliography{references}

@article{lueker1998exponentially,
  title={Exponentially small bounds on the expected optimum of the partition and subset sum problems},
  author={Lueker, George S},
  journal={Random Structures \& Algorithms},
  volume={12},
  number={1},
  pages={51--62},
  year={1998},
  publisher={Wiley Online Library}
}

@article{pensia2020optimal,
  title={Optimal lottery tickets via subset sum: Logarithmic over-parameterization is sufficient},
  author={Pensia, Ankit and Rajput, Shashank and Nagle, Alliot and Vishwakarma, Harit and Papailiopoulos, Dimitris},
  journal={Advances in neural information processing systems},
  volume={33},
  pages={2599--2610},
  year={2020}
}

@article{gao2017properties,
  title={On the properties of the softmax function with application in game theory and reinforcement learning},
  author={Gao, Bolin and Pavel, Lacra},
  journal={arXiv preprint arXiv:1704.00805},
  year={2017}
}

@article{vaswani2017attention,
  title={Attention is all you need},
  author={Vaswani, Ashish and Shazeer, Noam and Parmar, Niki and Uszkoreit, Jakob and Jones, Llion and Gomez, Aidan N and Kaiser, {\L}ukasz and Polosukhin, Illia},
  journal={Advances in neural information processing systems},
  volume={30},
  year={2017}
}

@inproceedings{devlin2019bert,
  title={{BERT}: Pre-training of deep bidirectional transformers for language understanding},
  author={Devlin, Jacob and Chang, Ming-Wei and Lee, Kenton and Toutanova, Kristina},
  booktitle={Proceedings of the 2019 conference of the North American chapter of the association for computational linguistics: human language technologies, volume 1 (long and short papers)},
  pages={4171--4186},
  year={2019}
}

@misc{gurobi,
  author = {{Gurobi Optimization, LLC}},
  title = {{Gurobi optimizer reference manual}},
  year = 2024,
  url = "https://www.gurobi.com"
}

@inproceedings{ramanujan2020s,
  title={What's hidden in a randomly weighted neural network?},
  author={Ramanujan, Vivek and Wortsman, Mitchell and Kembhavi, Aniruddha and Farhadi, Ali and Rastegari, Mohammad},
  booktitle={Proceedings of the IEEE/CVF conference on computer vision and pattern recognition},
  pages={11893--11902},
  year={2020}
}

@inproceedings{malach2020proving,
  title={Proving the lottery ticket hypothesis: Pruning is all you need},
  author={Malach, Eran and Yehudai, Gilad and Shalev-Schwartz, Shai and Shamir, Ohad},
  booktitle={International Conference on Machine Learning},
  pages={6682--6691},
  year={2020},
  organization={PMLR}
}

@article{orseau2020logarithmic,
  title={Logarithmic pruning is all you need},
  author={Orseau, Laurent and Hutter, Marcus and Rivasplata, Omar},
  journal={Advances in Neural Information Processing Systems},
  volume={33},
  pages={2925--2934},
  year={2020}
}

@inproceedings{da2022proving,
  title={Proving the strong lottery ticket hypothesis for convolutional neural networks},
  author={da Cunha, Arthur and Natale, Emanuele and Viennot, Laurent},
  booktitle={ICLR 2022-10th International Conference on Learning Representations},
  year={2022}
}

@article{da2023polynomially,
  title={Polynomially over-parameterized convolutional neural networks contain structured strong winning lottery tickets},
  author={Da Cunha, Arthur and d'Amore, Francesco},
  journal={Advances in Neural Information Processing Systems},
  volume={36},
  pages={25929--25957},
  year={2023}
}

@article{fischer2021towards,
  title={Towards strong pruning for lottery tickets with non-zero biases},
  author={Fischer, Jonas and Burkholz, Rebekka},
  journal={arXiv preprint arXiv:2110.11150},
  year={2021}
}

@inproceedings{burkholz2022convolutional,
  title={Convolutional and residual networks provably contain lottery tickets},
  author={Burkholz, Rebekka},
  booktitle={International Conference on Machine Learning},
  pages={2414--2433},
  year={2022},
  organization={PMLR}
}

@inproceedings{
ferbach2023a,
title={A general framework for proving the equivariant strong lottery ticket hypothesis},
author={Damien Ferbach and Christos Tsirigotis and Gauthier Gidel and Joey Bose},
booktitle={The Eleventh International Conference on Learning Representations },
year={2023},
url={https://openreview.net/forum?id=vVJZtlZB9D}
}

@article{
otsuka2025partially,
title={Partially frozen random networks contain compact strong lottery tickets},
author={Hikari Otsuka and Daiki Chijiwa and {\'A}ngel L{\'o}pez Garc{\'\i}a-Arias and Yasuyuki Okoshi and Kazushi Kawamura and Thiem Van Chu and Daichi Fujiki and Susumu Takeuchi and Masato Motomura},
journal={Transactions on Machine Learning Research},
issn={2835-8856},
year={2025},
url={https://openreview.net/forum?id=xpnPYfufhz},
note={}
}

@article{
ito2025uncovering,
title={Uncovering strong lottery tickets in graph transformers: A path to memory efficient and robust graph learning},
author={Hiroaki Ito and Jiale Yan and Hikari Otsuka and Kazushi Kawamura and Masato Motomura and Thiem Van Chu and Daichi Fujiki},
journal={Transactions on Machine Learning Research},
issn={2835-8856},
year={2025},
url={https://openreview.net/forum?id=B1q9po4LPl},
note={}
}

@inproceedings{shen-etal-2021-whats,
    title = "What`s hidden in a one-layer randomly weighted transformer?",
    author = "Shen, Sheng  and
      Yao, Zhewei  and
      Kiela, Douwe  and
      Keutzer, Kurt  and
      Mahoney, Michael",
    editor = "Moens, Marie-Francine  and
      Huang, Xuanjing  and
      Specia, Lucia  and
      Yih, Scott Wen-tau",
    booktitle = "Proceedings of the 2021 Conference on Empirical Methods in Natural Language Processing",
    month = nov,
    year = "2021",
    address = "Online and Punta Cana, Dominican Republic",
    publisher = "Association for Computational Linguistics",
    url = "https://aclanthology.org/2021.emnlp-main.231/",
    doi = "10.18653/v1/2021.emnlp-main.231",
    pages = "2914--2921"
}

@inproceedings{
frankle2018the,
title={The lottery ticket hypothesis: Finding sparse, trainable neural networks},
author={Jonathan Frankle and Michael Carbin},
booktitle={International Conference on Learning Representations},
year={2019},
url={https://openreview.net/forum?id=rJl-b3RcF7},
}

@article{zhou2019deconstructing,
  title={Deconstructing lottery tickets: Zeros, signs, and the supermask},
  author={Zhou, Hattie and Lan, Janice and Liu, Rosanne and Yosinski, Jason},
  journal={Advances in neural information processing systems},
  volume={32},
  year={2019}
}

@inproceedings{
diffenderfer2021multiprize,
title={Multi-prize lottery ticket hypothesis: Finding accurate binary neural networks by pruning a randomly weighted network},
author={James Diffenderfer and Bhavya Kailkhura},
booktitle={International Conference on Learning Representations},
year={2021},
url={https://openreview.net/forum?id=U_mat0b9iv}
}

@inproceedings{
burkholz2022most,
title={Most activation functions can win the lottery without excessive depth},
author={Rebekka Burkholz},
booktitle={Advances in Neural Information Processing Systems},
editor={Alice H. Oh and Alekh Agarwal and Danielle Belgrave and Kyunghyun Cho},
year={2022},
url={https://openreview.net/forum?id=NySDKS9SxN}
}

@inproceedings{gadhikar2023random,
  title={Why random pruning is all we need to start sparse},
  author={Gadhikar, Advait Harshal and Mukherjee, Sohom and Burkholz, Rebekka},
  booktitle={International Conference on Machine Learning},
  pages={10542--10570},
  year={2023},
  organization={PMLR}
}

@inproceedings{
natale2024on,
title={On the sparsity of the strong lottery ticket hypothesis},
author={Emanuele Natale and Davide Ferre' and Giordano Giambartolomei and Fr{\'e}d{\'e}ric Giroire and Frederik Mallmann-Trenn},
booktitle={The Thirty-eighth Annual Conference on Neural Information Processing Systems},
year={2024},
url={https://openreview.net/forum?id=aBMESB1Ajx}
}

@inproceedings{shen-etal-2021-reservoir,
    title = "Reservoir transformers",
    author = "Shen, Sheng  and
      Baevski, Alexei  and
      Morcos, Ari  and
      Keutzer, Kurt  and
      Auli, Michael  and
      Kiela, Douwe",
    editor = "Zong, Chengqing  and
      Xia, Fei  and
      Li, Wenjie  and
      Navigli, Roberto",
    booktitle = "Proceedings of the 59th Annual Meeting of the Association for Computational Linguistics and the 11th International Joint Conference on Natural Language Processing (Volume 1: Long Papers)",
    month = aug,
    year = "2021",
    address = "Online",
    publisher = "Association for Computational Linguistics",
    url = "https://aclanthology.org/2021.acl-long.331/",
    doi = "10.18653/v1/2021.acl-long.331",
    pages = "4294--4309"
}

@inproceedings{
zhong2024algorithmic,
title={Algorithmic capabilities of random transformers},
author={Ziqian Zhong and Jacob Andreas},
booktitle={The Thirty-eighth Annual Conference on Neural Information Processing Systems},
year={2024},
url={https://openreview.net/forum?id=plH8gW7tPQ}
}

@inproceedings{
merity2017pointer,
title={Pointer sentinel mixture models},
author={Stephen Merity and Caiming Xiong and James Bradbury and Richard Socher},
booktitle={International Conference on Learning Representations},
year={2017},
url={https://openreview.net/forum?id=Byj72udxe}
}

@inproceedings{
loshchilov2018decoupled,
title={Decoupled weight decay regularization},
author={Ilya Loshchilov and Frank Hutter},
booktitle={International Conference on Learning Representations},
year={2019},
url={https://openreview.net/forum?id=Bkg6RiCqY7},
}

@article{radford2019language,
  title={Language models are unsupervised multitask learners},
  author={Radford, Alec and Wu, Jeffrey and Child, Rewon and Luan, David and Amodei, Dario and Sutskever, Ilya and others},
  journal={OpenAI blog},
  volume={1},
  number={8},
  pages={9},
  year={2019}
}

@ARTICLE{2020SciPy-NMeth,
  author  = {Virtanen, Pauli and Gommers, Ralf and Oliphant, Travis E. and
            Haberland, Matt and Reddy, Tyler and Cournapeau, David and
            Burovski, Evgeni and Peterson, Pearu and Weckesser, Warren and
            Bright, Jonathan and {van der Walt}, St{\'e}fan J. and
            Brett, Matthew and Wilson, Joshua and Millman, K. Jarrod and
            Mayorov, Nikolay and Nelson, Andrew R. J. and Jones, Eric and
            Kern, Robert and Larson, Eric and Carey, C J and
            Polat, {\.I}lhan and Feng, Yu and Moore, Eric W. and
            {VanderPlas}, Jake and Laxalde, Denis and Perktold, Josef and
            Cimrman, Robert and Henriksen, Ian and Quintero, E. A. and
            Harris, Charles R. and Archibald, Anne M. and
            Ribeiro, Ant{\^o}nio H. and Pedregosa, Fabian and
            {van Mulbregt}, Paul and {SciPy 1.0 Contributors}},
  title   = {{{SciPy} 1.0: Fundamental algorithms for scientific
            computing in python}},
  journal = {Nature Methods},
  year    = {2020},
  volume  = {17},
  pages   = {261--272},
  adsurl  = {https://rdcu.be/b08Wh},
  doi     = {10.1038/s41592-019-0686-2},
}

@article{chijiwa2021pruning,
  title={Pruning randomly initialized neural networks with iterative randomization},
  author={Chijiwa, Daiki and Yamaguchi, Shin'ya and Ida, Yasutoshi and Umakoshi, Kenji and Inoue, Tomohiro},
  journal={Advances in neural information processing systems},
  volume={34},
  pages={4503--4513},
  year={2021}
}

@inproceedings{xiong2023strong,
  title={Strong lottery ticket hypothesis with $\varepsilon$-perturbation},
  author={Xiong, Zheyang and Liao, Fangshuo and Kyrillidis, Anastasios},
  booktitle={International Conference on Artificial Intelligence and Statistics},
  pages={6879--6902},
  year={2023},
  organization={PMLR}
}

@InProceedings{pmlr-v9-glorot10a,
  title = 	 {Understanding the difficulty of training deep feedforward neural networks},
  author = 	 {Glorot, Xavier and Bengio, Yoshua},
  booktitle = 	 {Proceedings of the Thirteenth International Conference on Artificial Intelligence and Statistics},
  pages = 	 {249--256},
  year = 	 {2010},
  editor = 	 {Teh, Yee Whye and Titterington, Mike},
  volume = 	 {9},
  series = 	 {Proceedings of Machine Learning Research},
  address = 	 {Chia Laguna Resort, Sardinia, Italy},
  month = 	 {13--15 May},
  publisher =    {PMLR},
  url = 	 {https://proceedings.mlr.press/v9/glorot10a.html}
}

@inproceedings{
loshchilov2017sgdr,
title={{SGDR}: Stochastic gradient descent with warm restarts},
author={Ilya Loshchilov and Frank Hutter},
booktitle={International Conference on Learning Representations},
year={2017},
url={https://openreview.net/forum?id=Skq89Scxx}
}

@inproceedings{wolf-etal-2020-transformers,
    title = "HuggingFace's transformers: State-of-the-art natural language processing",
    author = "Wolf, Thomas  and
      Debut, Lysandre  and
      Sanh, Victor  and
      Chaumond, Julien  and
      Delangue, Clement  and
      Moi, Anthony  and
      Cistac, Pierric  and
      Rault, Tim  and
      Louf, Remi  and
      Funtowicz, Morgan  and
      Davison, Joe  and
      Shleifer, Sam  and
      von Platen, Patrick  and
      Ma, Clara  and
      Jernite, Yacine  and
      Plu, Julien  and
      Xu, Canwen  and
      Le Scao, Teven  and
      Gugger, Sylvain  and
      Drame, Mariama  and
      Lhoest, Quentin  and
      Rush, Alexander",
    editor = "Liu, Qun  and
      Schlangen, David",
    booktitle = "Proceedings of the 2020 Conference on Empirical Methods in Natural Language Processing: System Demonstrations",
    month = oct,
    year = "2020",
    address = "Online",
    publisher = "Association for Computational Linguistics",
    url = "https://aclanthology.org/2020.emnlp-demos.6/",
    doi = "10.18653/v1/2020.emnlp-demos.6",
    pages = "38--45"
}

\newpage
\onecolumn
\appendix
\section{Proofs of Main Theorems}\label{sec:proofs_app}
    This section presents the detailed proofs of the main theorems in the manuscript.
    We first introduce two lemmas: 
    one for approximating a target weight matrix by pruning two random weight matrices, and another for bounding the effect of perturbations in the softmax function. 
    These lemmas are then used to establish the SLTH for attention mechanisms.
    Then, leveraging the theory of the SLTH for attention mechanisms, we prove the existence of SLTs in transformer blocks and transformers without normalization layers.
    
    \subsection{Weight Approximation}\label{subsec:weight_approximation_app}
        \citet{pensia2020optimal} have shown that a two-layer fully-connected ReLU network can approximate arbitrary matrices with high probability.
        Our problem setting can be viewed as a simplified version of their construction, in which the ReLU nonlinearity is omitted.
        We follow their proof strategy and simplify it to the linear (non-activated) case.
        
        \begin{lemma}\label{lem:weight_approximation_app}
            Let ${\bm{W} \in \mathbb{R}^{d_2 \times d_1}}$ be a target matrix with entries in ${[-1, 1]}$.
            Let ${\tilde{\bm{W}}_1 \in \mathbb{R}^{n \times d_1}}$ and ${\tilde{\bm{W}}_2 \in \mathbb{R}^{d_2 \times n}}$ be source random matrices whose entries are drawn i.i.d. from $U[-1, 1]$.
            For any $0 < \epsilon < 1$, suppose that $n \ge d_1 C \log( \frac{d_1 d_2}{\epsilon})$ for some universal constant $C > 0$.
            Then, with probability at least $1 - \epsilon$, there exists a choice of binary masks ${\bm{M}_1 \in \{0, 1\}^{n \times d_1}}$ and ${\bm{M}_2 \in \{0, 1\}^{d_2 \times n}}$ such that
            \begin{align}
                \nonumber
                \left\| \bm{W} - (\tilde{\bm{W}}_2 \odot \bm{M}_2)(\tilde{\bm{W}}_1 \odot \bm{M}_1) \right\|_{\max} \le \frac{\epsilon}{d_1 d_2}.
            \end{align}
            \begin{proof}
                Firstly, we structurally prune the random weight matrix $\tilde{\bm{W}}_1$ by the pruning mask $\bm{M}_1$:
                \begin{align}
                    \label{eq:right_side}
                    \tilde{\bm{W}}_1 \odot \bm{M}_1 =
                    \begin{bmatrix}
                    \bm{u}_1 & 0 & \cdots & 0 \\
                    0 & \bm{u}_2 & \cdots & 0 \\
                    \vdots & \vdots & \ddots & \vdots \\
                    0 & 0 & \cdots & \bm{u}_{d_1}
                    \end{bmatrix},
                \end{align}
                where $\bm{u}_i \in \mathbb{R}^{n'}$.
                Next, we decompose $\tilde{\bm{W}}_2 \odot \bm{M}_2$ as follows:
                \begin{align}
                    \label{eq:left_side}
                    \tilde{\bm{W}}_2 \odot \bm{M}_2 =
                    \begin{bmatrix}
                    \left(\bm{v}_{1,1} \odot \bm{m}_{1,1}\right)^\top & \left(\bm{v}_{1,2} \odot \bm{m}_{1,2} \right)^\top& \cdots & \left(\bm{v}_{1,d_1} \odot \bm{m}_{1,d_1}\right)^\top \\
                    \left(\bm{v}_{2,1} \odot \bm{m}_{2,1} \right)^\top& \left(\bm{v}_{2,2} \odot \bm{m}_{2,2} \right)^\top& \cdots & \left(\bm{v}_{2,d_1} \odot \bm{m}_{2,d_1} \right)^\top\\
                    \vdots & \vdots & \ddots & \vdots \\
                    \left(\bm{v}_{d_2,1} \odot \bm{m}_{d_2,1} \right)^\top& \left(\bm{v}_{d_2,2} \odot \bm{m}_{d_2,2} \right)^\top& \cdots & \left(\bm{v}_{d_2,d_1} \odot \bm{m}_{d_2,d_1} \right)^\top
                    \end{bmatrix},
                \end{align}
                where $\bm{v}_{i,j} \in \mathbb{R}^{n'}$ and $\bm{m}_{i,j} \in \{0, 1\}^{n'}$.
                These operations enable us to rewrite the product of \Cref{eq:right_side,eq:left_side} as follows:
                \begin{align}
                    \label{eq:left_right_matmul}
                    (\tilde{\bm{W}}_2 \odot \bm{M}_2)(\tilde{\bm{W}}_1 \odot \bm{M}_1) 
                    &=
                    \begin{bmatrix}
                    \left(\bm{v}_{1,1} \odot \bm{m}_{1,1}\right)^\top \bm{u}_1 & \left(\bm{v}_{1,2} \odot \bm{m}_{1,2} \right)^\top  \bm{u}_2 & \cdots & \left(\bm{v}_{1,d_1} \odot \bm{m}_{1,d_1}\right)^\top \bm{u}_{d_1}  \\
                    \left(\bm{v}_{2,1} \odot \bm{m}_{2,1} \right)^\top  \bm{u}_1 & \left(\bm{v}_{2,2} \odot \bm{m}_{2,2} \right)^\top \bm{u}_2 & \cdots & \left(\bm{v}_{2,d_1} \odot \bm{m}_{2,d_1} \right)^\top \bm{u}_{d_1} \\
                    \vdots & \vdots & \ddots & \vdots \\
                    \left(\bm{v}_{d_2,1} \odot \bm{m}_{d_2,1} \right)^\top  \bm{u}_1 & \left(\bm{v}_{d_2,2} \odot \bm{m}_{d_2,2} \right)^\top \bm{u}_2 & \cdots & \left(\bm{v}_{d_2,d_1} \odot \bm{m}_{d_2,d_1} \right)^\top \bm{u}_{d_1}
                    \end{bmatrix}
                \end{align}
                We focus on the $(i, j)$-th entry of \Cref{eq:left_right_matmul}.
                This entry can be rewritten as a subset sum of element-wise products between the vectors $\bm{v}_{i,j}$ and $\bm{u}_j$:
                \begin{align}
                    \nonumber
                    \left(\bm{v}_{i,j} \odot \bm{m}_{i,j}\right)^\top \bm{u}_j
                    = 
                    \sum_{k=1}^{n'} m_{i,j,k}v_{i,j,k}u_{j,k}.
                \end{align}
                Here, each $m_{i,j,k}$ determines whether the corresponding product $v_{i,j,k} u_{j,k}$ is included in the subset sum.
                We aim to approximate the ($i,j$)-th entry of the target weight matrix $\bm{W}$ with the subset sum $\sum_{k=1}^{n'} m_{i,j,k}v_{i,j,k}u_{j,k}$ by appropriately choosing the binary mask $\bm{m}_{i,j}$. 
                Since each entry of $v_{i,j,k}$ and $u_{j,k}$ is drawn i.i.d. from $U[-1, 1]$, each product $v_{i,j,k} u_{j,k}$ can be viewed as drawn from the distribution including some uniform distribution; thus, we can apply Corollary~3.3 of \citet{lueker1998exponentially}, 
                which states that if $n' \ge C \log \left( \frac{d_1 d_2}{\epsilon} \right)$, then with probability at least $1 - \frac{\epsilon}{d_1 d_2}$, there exists a binary mask vector $\bm{m}_{i,j}$ such that the subset sum $\sum_{k=1}^{n'} m_{i,j,k}v_{i,j,k}u_{j,k}$ approximates the ($i,j$)-th entry of $\bm{W}$ within an error of $\frac{\epsilon}{d_1 d_2}$.
                By the union bound, the probability that all entries of the weight matrix $\bm{W}$ are simultaneously approximated is at least $1-\epsilon$:
                \begin{align}
                    \nonumber
                    1 - \sum_{i=1}^{d_2}\sum_{j=1}^{d_1}\frac{\epsilon}{d_1d_2}=1-\epsilon.
                \end{align}
                Therefore, if $n = d_1 n' \ge d_1 C \log \left( \frac{d_1 d_2}{\epsilon} \right)$, then with probability at least $1 - \epsilon$, the following inequality holds:
                \begin{align}
                    \nonumber
                    \left\| \bm{W} - (\tilde{\bm{W}}_2 \odot \bm{M}_2)(\tilde{\bm{W}}_1 \odot \bm{M}_1) \right\|_{\max} \le \frac{\epsilon}{d_1 d_2}.
                \end{align}
            \end{proof}
        \end{lemma}

    \subsection{Spectral Norm of Softmax Difference}
        In addition to approximating target weights, we need to analyze the stability of the softmax output under small input perturbations, with respect to the spectral norm of the resulting attention-weighted output.

        \begin{lemma}\label{lem:softmax_diff_bound_app}
            Given ${\bm{\epsilon} \in \mathbb{R}^{d_1}}$ as a perturbation vector such that ${\|\bm{\epsilon}\|_{\max} \leq \epsilon_{\max}}$ for some ${\epsilon_{\max} \ge 0}$, we have 
            \begin{align*}
                \left\| \softmax(\bm{x}_i; \bm{a}_i) \bm{X} - \softmax(\bm{x}_i + \bm{\epsilon}; \bm{a}_i) \bm{X} \right\| \leq 
                \sqrt{d_1} \alpha \left( \exp(2 \epsilon_{\max}) - 1 \right).
            \end{align*}
            \begin{proof}
            Let $\bm{p}_i := \softmax(\bm{x}_i; \bm{a}_i)$ and $\bm{p}_i' := \softmax(\bm{x}_i + \bm{\epsilon}; \bm{a}_i)$. 
            Then, for each coordinate $j$, we have
            \begin{align*}
                p'_{i,j} &= p_{i,j} \frac{\exp(\epsilon_j)}{Z}, \\
                Z &= \sum_{k=1}^T p_{i, k} \exp(\epsilon_k).
            \end{align*}
            By the assumption ${\|\bm{\epsilon}\|_{\max} \le \epsilon_{\max}}$, we have the following bound:
            \begin{align}
                \label{eq:exp_esp_minus_one}
                \left| 1 - \frac{\exp(\epsilon_j)}{Z} \right| \le \exp(2 \epsilon_{\max}) - 1.
            \end{align}
            Now, we can bound the spectral norm for the $i$-th input embedding:
            \begin{align*}
                \left\| \bm{p}_i \bm{X} - \bm{p}'_i \bm{X} \right\|
                &\le \sqrt{d_1} \cdot \left\| \bm{p}_i \bm{X} - \bm{p}'_i \bm{X} \right\|_{\max} \\
                &\le \sqrt{d_1} \cdot \max_{j \in [d_1]} \left| \sum_{k=1}^T (p_{i,k} - p'_{i,k}) x_{k, j} \right| \\
                &\le \sqrt{d_1} \cdot \max_{j \in [d_1]} \sum_{k=1}^T |x_{k, j}| \cdot |p_{i,k} - p'_{i,k}| \\
                &\le \sqrt{d_1} \cdot \alpha \sum_{k=1}^T |p_{i,k} - p'_{i,k}| \\
                &\le \sqrt{d_1} \cdot \alpha \sum_{k=1}^T p_{i,k} \left| 1 - \frac{\exp(\epsilon_k)}{Z} \right| \\
                &\le \sqrt{d_1} \cdot \alpha \left( \exp(2 \epsilon_{\max}) - 1 \right) \sum_{k=1}^T p_{i,k} \tag{Using \Cref{eq:exp_esp_minus_one}} \\
                &= \sqrt{d_1} \cdot \alpha \left( \exp(2 \epsilon_{\max}) - 1 \right).
            \end{align*}
            This upper bound is independent of $i$; thus, the upper bound of $\max_{i\in[T]}\left\| \bm{p} \bm{X} - \bm{p}' \bm{X} \right\|$ is same as the final upper bound.
            \end{proof}
        \end{lemma}
    
        \subsection{SLT Existence within Attention Mechanisms}\label{subsec:SLT_existence_within_attention_mechanisms_app}
        By leveraging these two lemmas, we prove the following theorem:
        \begin{theorem}\label{thr:slt_existence_attn_mech_app}
            Let $\sattn(\cdot)$ and $\tattn(\cdot)$ be as defined in \Cref{eq:def_sattn,eq:def_tattn}.
            Assume $\alpha \ge \max(\sqrt{d_1}, \sqrt{d_2})$ for the inputs.
            Then, with probability at least $1 - \epsilon$, there exists a choice of binary masks $\bm{M}_{\mathrm Q}^{(j)}, \bm{M}_{\mathrm K}^{(j)}, \bm{M}_{\mathrm V}^{(j)}, \bm{M}_{\mathrm O}^{(j)}$ that satisfy
            \[
                \max_{i \in [T]} \|\sattn(\bm{x}_i) - \tattn(\bm{x}_i)\| \le \epsilon,
            \]
            if the source dimensions satisfy
            \begin{align*}
                n_{1} &\ge d_1 C \log\left(\frac{8H\alpha^3 d_1^{3/2}}{\epsilon}\right), \\
                n_{2} &\ge d_1 C \log\left(\frac{2H\alpha d_1 \sqrt{d_2}}{\epsilon}\right),
            \end{align*}
            for some universal constant $C > 0$.
            \begin{proof}
                We prove the theorem in three steps.
                \paragraph{Step 1: Weight Merging.}
                We begin by merging the weight matrices of the target and source MHAs.
                The target MHA weights are merged as
                \begin{align}
                    \label{eq:w_qk}
                    \bm{W}_{\mathrm{QK}}^{(j)} &:= \frac{1}{\sqrt{d_{\textrm{K}}}}\bm{W}_{\mathrm Q}^{(j)}(\bm{W}_{\mathrm K}^{(j)})^\top, \\
                    \label{eq:w_vo}
                    \bm{W}_{\mathrm{VO}}^{(j)} &:= \bm{W}_{\mathrm V}^{(j)}\bm{W}_{\mathrm O}^{(j)}.
                \end{align}
                This operation (\Cref{eq:w_qk,eq:w_vo}) enables us to represent each head of $\tattn(\cdot)$ as
                \begin{align*}
                    \tattn(\bm{x}_i; \bm{X}, \bm{W}_{\mathrm{Q:O}}^{(1:H)}) 
                    = \sum_{j=1}^H \softmax\left(\bm{x}_i^\top \bm{W}_{\mathrm{QK}}^{(j)} \bm{X}^\top; \bm{a}_i\right)\bm{X} \bm{W}_{\mathrm{VO}}^{(j)}.
                \end{align*}
                From the assumption on the target weights, we have the following norm bounds:
                \begin{align}
                    \label{eq:w_qk_norm}
                    \|\bm{W}_{\mathrm{QK}}^{(j)}\| &\le 1/\sqrt{d_{\textrm{K}}}, \\
                    \label{eq:w_vo_norm}
                    \|\bm{W}_{\mathrm{VO}}^{(j)}\| &\le 1.
                \end{align}
                For the source MHA, we incorporate the scaling factor $1/\sqrt{n_{\textrm{K}}}$ into the query and key weight matrices:
                \begin{align*}
                    \tilde{\bm{W}}_{\mathrm{Q}}'^{(j)} &:= \frac{1}{d_{\textrm{K}}^{1/4}}\tilde{\bm{W}}_{\mathrm Q}^{(j)}, \\
                    \tilde{\bm{W}}_{\mathrm{K}}'^{(j)} &:= \frac{1}{d_{\textrm{K}}^{1/4}}\tilde{\bm{W}}_{\mathrm K}^{(j)}.
                \end{align*}
                Assuming that each entry of $\tilde{\bm{W}}_{\mathrm{Q}}^{(j)}$ and $\tilde{\bm{W}}_{\mathrm{K}}^{(j)}$ is drawn i.i.d. from $U[-d_{\textrm{K}}^{1/4}, d_{\textrm{K}}^{1/4}]$, each entry of the scaled matrices $\tilde{\bm{W}}_{\mathrm{Q}}'^{(j)}$ and $\tilde{\bm{W}}_{\mathrm{K}}'^{(j)}$ is drawn i.i.d. from $U[-1, 1]$.
                
                \paragraph{Step 2: Weight Approximation.}
                From \Cref{lem:weight_approximation_app}, for any $0 < \epsilon < 1$, if 
                \[
                    n_{\mathrm{K}} \ge d_1 C \log\left(\frac{8H\alpha^3 d_1^{3/2}}{\epsilon}\right),
                \]
                then with probability at least $1 - \frac{\epsilon}{8H\alpha^3\sqrt{d_1}}$, there exists a choice of binary masks $\bm{M}_{\mathrm Q}^{(j)}$ and $\bm{M}_{\mathrm K}^{(j)}$ such that
                \begin{align}
                    \label{eq:w_qk_bound}
                    \left\| \bm{W}_{\mathrm{QK}}^{(j)} - \left(\tilde{\bm{W}}_{\mathrm Q}'^{(j)} \odot \bm{M}_{\mathrm Q}^{(j)}\right)\left(\tilde{\bm{W}}_{\mathrm K}'^{(j)} \odot \bm{M}_{\mathrm K}^{(j)}\right)^\top \right\|_{\max}
                    \le 
                    \frac{\epsilon}{8H\alpha^3 d_1^{3/2}}.
                \end{align}
                This inequality \Cref{eq:w_qk_bound} implies a bound on the softmax input:
                \begin{align*}
                    &\left\|\bm{x}_i^\top \bm{W}_{\mathrm{QK}}^{(j)} \bm{X}^\top - \bm{x}_i^\top \left(\tilde{\bm{W}}_{\mathrm Q}'^{(j)} \odot \bm{M}_{\mathrm Q}^{(j)}\right) \left(\tilde{\bm{W}}_{\mathrm K}'^{(j)} \odot \bm{M}_{\mathrm K}^{(j)}\right)^\top \bm{X}^\top \right\|_\infty \\
                    &= 
                    \max_{k \in [T]} \left| \bm{x}_i^\top \bm{W}_{\mathrm{QK}}^{(j)} \bm{x}_k - \bm{x}_i^\top \left(\tilde{\bm{W}}_{\mathrm Q}'^{(j)} \odot \bm{M}_{\mathrm Q}^{(j)}\right) \left(\tilde{\bm{W}}_{\mathrm K}'^{(j)} \odot \bm{M}_{\mathrm K}^{(j)}\right)^\top \bm{x}_k \right| \\
                    &\le 
                    \alpha^2 \left\| \bm{W}_{\mathrm{QK}}^{(j)} - \left(\tilde{\bm{W}}_{\mathrm Q}'^{(j)} \odot \bm{M}_{\mathrm Q}^{(j)}\right) \left(\tilde{\bm{W}}_{\mathrm K}'^{(j)} \odot \bm{M}_{\mathrm K}^{(j)}\right)^\top \right\| \\
                    &\le 
                    \alpha^2 d_1 \left\| \bm{W}_{\mathrm{QK}}^{(j)} - \left(\tilde{\bm{W}}_{\mathrm Q}'^{(j)} \odot \bm{M}_{\mathrm Q}^{(j)}\right) \left(\tilde{\bm{W}}_{\mathrm K}'^{(j)} \odot \bm{M}_{\mathrm K}^{(j)}\right)^\top \right\|_{\max} \\
                    &\le 
                    \alpha^2 d_1 \frac{\epsilon}{8H\alpha^3 d_1^{3/2}} \\
                    &= 
                    \frac{\epsilon}{8H\alpha \sqrt{d_1}}.
                \end{align*}
                Let
                \begin{align*}
                    \bm{p}_i^{(j)} 
                    &:= 
                    \softmax\left(\bm{x}_i^\top \bm{W}_{\mathrm{QK}}^{(j)} \bm{X}^\top ; \bm{a}_i\right), \\
                    \bm{p}_i'^{(j)} 
                    &:= 
                    \softmax\left(\bm{x}_i^\top (\tilde{\bm{W}}_{\mathrm Q}'^{(j)} \odot \bm{M}_{\mathrm Q}^{(j)})(\tilde{\bm{W}}_{\mathrm K}'^{(j)} \odot \bm{M}_{\mathrm K}^{(j)})^\top \bm{X}^\top; \bm{a}_i\right).
                \end{align*}
                Applying \Cref{lem:softmax_diff_bound_app}, we obtain
                \begin{align*}
                    \left\|\bm{p}_i^{(j)}\bm{X} - \bm{p}_i'^{(j)}\bm{X} \right\| 
                    &\le 
                    \sqrt{d_1}\alpha\left(\exp\left(\frac{\epsilon}{4H\alpha\sqrt{d_1}}\right) - 1\right) \\
                    &\le 
                    \frac{\epsilon}{2H}. \tag{Using $0 < \frac{\epsilon}{4H\alpha\sqrt{d_1}} < 1$}
                \end{align*}
                For the value and output weights, from \Cref{lem:weight_approximation_app}, if
                \[
                    n_{\mathrm{V}} \ge d_1 C \log\left(\frac{2H\alpha d_1 \sqrt{d_2}}{\epsilon_2}\right),
                \]
                then with probability at least $1 - \frac{\sqrt{d_2} \epsilon}{2H\alpha}$, there exists a choice of binary pruning masks $\bm{M}_{\mathrm V}^{(j)}$ and $\bm{M}_{\mathrm O}^{(j)}$ such that
                \begin{align}
                    \label{eq:w_vo_bound}
                    \left\|\bm{W}_{\mathrm{VO}}^{(j)} - (\tilde{\bm{W}}_{\mathrm V}^{(j)} \odot \bm{M}_{\mathrm V}^{(j)})(\tilde{\bm{W}}_{\mathrm O}^{(j)} \odot \bm{M}_{\mathrm O}^{(j)}) \right\|_{\max} 
                    \le 
                    \frac{\epsilon}{2H\alpha d_1 \sqrt{d_2}}.
                \end{align}
                
                \paragraph{Step 3: Total Error Analysis.}
                We now bound the difference between the outputs of the source and target MHAs:
                \begin{align*}
                    \left\|\tattn(\bm{x}_i) - \sattn(\bm{x}_i)\right\|
                    &= 
                    \left\| \sum_{j=1}^H \left(\bm{p}_i^{(j)} \bm{X} \bm{W}_{\mathrm{VO}}^{(j)} - \bm{p}_i'^{(j)} \bm{X} (\tilde{\bm{W}}_{\mathrm V}^{(j)} \odot \bm{M}_{\mathrm V}^{(j)})(\tilde{\bm{W}}_{\mathrm O}^{(j)} \odot \bm{M}_{\mathrm O}^{(j)})\right) \right\| \\
                    &\le
                    \sum_{j=1}^H \left\| \left(\bm{p}_i^{(j)} \bm{X} \bm{W}_{\mathrm{VO}}^{(j)} - \bm{p}_i'^{(j)} \bm{X} (\tilde{\bm{W}}_{\mathrm V}^{(j)} \odot \bm{M}_{\mathrm V}^{(j)})(\tilde{\bm{W}}_{\mathrm O}^{(j)} \odot \bm{M}_{\mathrm O}^{(j)})\right) \right\|.
                \end{align*}
                We apply the triangle inequality:
                \begin{align*}
                    &\left\|\bm{p}_i^{(j)} \bm{X} \bm{W}_{\mathrm{VO}}^{(j)} - \bm{p}_i'^{(j)} \bm{X} (\tilde{\bm{W}}_{\mathrm V}^{(j)} \odot \bm{M}_{\mathrm V}^{(j)})(\tilde{\bm{W}}_{\mathrm O}^{(j)} \odot \bm{M}_{\mathrm O}^{(j)}) \right\| \\
                    &\le
                    \left\| (\bm{p}_i^{(j)} - \bm{p}_i'^{(j)}) \bm{X} \bm{W}_{\mathrm{VO}}^{(j)} \right\| + \left\| \bm{p}_i'^{(j)} \bm{X} (\bm{W}_{\mathrm{VO}}^{(j)} - (\tilde{\bm{W}}_{\mathrm V}^{(j)} \odot \bm{M}_{\mathrm V}^{(j)})(\tilde{\bm{W}}_{\mathrm O}^{(j)} \odot \bm{M}_{\mathrm O}^{(j)})) \right\|.
                \end{align*}
                For the first term, by using $\|\bm{W}_{\mathrm{VO}}^{(j)}\| \le 1$ (\Cref{eq:w_vo_norm}), we obtain
                \begin{align}
                    \nonumber
                    \left\| (\bm{p}_i^{(j)} - \bm{p}_i'^{(j)}) \bm{X} \bm{W}_{\mathrm{VO}}^{(j)} \right\|
                    &\le 
                     \left\| (\bm{p}_i^{(j)} - \bm{p}_i'^{(j)}) \bm{X}\right\|\left\| \bm{W}_{\mathrm{VO}}^{(j)} \right\| \\ 
                     \nonumber
                    &\le 
                    \left\| (\bm{p}_i^{(j)} - \bm{p}_i'^{(j)}) \bm{X}\right\| \\
                    \label{eq:first_term_slt_existence_attn}
                    &\le 
                    \frac{\epsilon}{2H}.
                \end{align}
                For the second term, we obtain the following result by using \Cref{eq:w_vo_bound}:
                \begin{align}
                    \nonumber
                    &\left\| \bm{p}_i'^{(j)} \bm{X} (\bm{W}_{\mathrm{VO}}^{(j)} - (\tilde{\bm{W}}_{\mathrm V}^{(j)} \odot \bm{M}_{\mathrm V}^{(j)})(\tilde{\bm{W}}_{\mathrm O}^{(j)} \odot \bm{M}_{\mathrm O}^{(j)})) \right\| \\
                    \nonumber
                    &\le
                    \sqrt{d_1}\left\| \bm{p}_i'^{(j)} \bm{X} \right\|_{\infty} \sqrt{d_1 d_2}\left\|\bm{W}_{\mathrm{VO}}^{(j)} - (\tilde{\bm{W}}_{\mathrm V}^{(j)} \odot \bm{M}_{\mathrm V}^{(j)})(\tilde{\bm{W}}_{\mathrm O}^{(j)} \odot \bm{M}_{\mathrm O}^{(j)}) \right\|_{\max} \\
                    \nonumber
                    &\le
                    d_1\sqrt{d_2}\alpha \left\|\bm{W}_{\mathrm{VO}}^{(j)} - (\tilde{\bm{W}}_{\mathrm V}^{(j)} \odot \bm{M}_{\mathrm V}^{(j)})(\tilde{\bm{W}}_{\mathrm O}^{(j)} \odot \bm{M}_{\mathrm O}^{(j)}) \right\|_{\max} \\
                    \nonumber
                    &\le
                    d_1\sqrt{d_2}\alpha \frac{\epsilon}{2H\alpha d_1 \sqrt{d_2}} \tag{Using \Cref{eq:w_vo_bound}} \\
                    \label{eq:second_term_slt_existence_attn}
                    &=
                    \frac{\epsilon}{2H}.
                \end{align}
                These results of \Cref{eq:first_term_slt_existence_attn,eq:second_term_slt_existence_attn} do not depend on the input index $i$; 
                thus, adding the two terms across $H$ heads gives
                \begin{align*}
                    \max_{i \in [T]} \left\|\tattn(\bm{x}_i) - \sattn(\bm{x}_i)\right\|
                    &\le 
                    \sum_{j=1}^H \left(\frac{\epsilon}{2H} + \frac{\epsilon}{2H}\right) 
                    =
                    \epsilon.
                \end{align*}
                Finally, using the union bound and the assumption $\alpha\ge \max(\sqrt{d_1}, \sqrt{d_2})$, the probability that all approximations hold is
                \begin{align*}
                    1 - \frac{\epsilon}{8H\alpha^3 \sqrt{d_1}} - \frac{\sqrt{d_2}\epsilon}{2H\alpha} 
                    \ge
                    1 - \epsilon.
                \end{align*}
            \end{proof}
        \end{theorem}

        \subsection{SLT Existence Within Transformer Blocks}
            
            By combining the SLT existence theorem for attention mechanisms (\Cref{thr:slt_existence_attn_mech_app}) and for multi-layer fully-connected ReLU networks (FC) proven by~\citet{pensia2020optimal} (\Cref{thr:slt_existence_within_ffn_app}), we prove the SLT existence theorem for transformer blocks.
            \begin{theorem}[Theorem~1 in \citet{pensia2020optimal}]\label{thr:slt_existence_within_ffn_app}
                Let 
                \[
                    \tffn\left(\bm{x}_i\right) = \bm{W}_L\relu(\bm{W}_{L-1}\dots\relu(\bm{W}_1\bm{x}_i))
                \]
                be a target FC with $L$ layers. 
                Assume that each weight matrix $\bm{W}_l \in \mathbb{R}^{d_{l+1} \times d_{l}}$ satisfies $|\bm{W}_l| \le 1$ for all $l = 1, \dots, L$.
                Consider a pruned source FC with $2L$ layers defined as
                \begin{align*}
                    \sffn\left(\bm{x}_i\right) 
                    &= \left(\tilde{\bm{W}}_{2L} \odot \bm{M}_{2L}\right)\relu\left(\left(\tilde{\bm{W}}_{2L-1} \odot \bm{M}_{2L-1}\right) \dots \relu\left(\left(\tilde{\bm{W}}_{2L-1} \odot \bm{M}_{2L-1}\right)\bm{x}_i\right)\right),
                \end{align*}
                where $\tilde{\bm{W}}_{2l-1} \in \mathbb{R}^{n_l\times d_l}$ and $\tilde{\bm{W}}_{2l} \in \mathbb{R}^{d_{l+1}\times n_{l}}$ for ${l=1, \dots, L}$,
                and each entry of $\tilde{\bm{W}}_{l}$ is drawn i.i.d. from $U[-1, 1]$.
                Then, with probability at least $1 - \epsilon$ for any ${0 < \epsilon < 1}$, there exists a choice of binary pruning masks $\bm{M}_1, \dots, \bm{M}_{2L}$ that holds the following inequality:
                \begin{align*}
                    \left\| \sffn(\bm{x}_i) - \tffn(\bm{x}_i) \right\| 
                    \le 
                    \exp\left(\frac{\alpha \epsilon}{2}\right)-1,
                \end{align*}
                if each source dimension $n_{l}$ satisfies
                \begin{align*}
                    n_{l} \ge d_l C\log\frac{4 L d_l d_{l+1}}{\epsilon},
                \end{align*}
                for some universal constant $C>0$.
            \end{theorem}
            We now state the main result for transformer blocks, which follows from combining the two SLT existence theorems.
             \begin{theorem}\label{thr:slt_existence_within_attn_block_app}
                Let 
                \[
                    \tblk (\bm{x}_i)=\tffn(\tattn(\bm{x}_i)^\top + \bm{x}_i) + \tattn(\bm{x}_i)^\top + \bm{x}_i
                \]
                be a target transformer block of an MHA $\tattn(\cdot)$ and FC $\tffn(\cdot)$ with $L$ layers.
                For simplicity, we assume each layer of target FC dimensions is all $d_1$.
                Let 
                \[
                    \sblk (\bm{x}_i)=\sffn(\sattn(\bm{x}_i)^\top + \bm{x}_i) + \sattn(\bm{x}_i)^\top + \bm{x}_i
                \]
                be a pruned random source transformer block of a pruned random MHA $\sattn(\cdot)$ and FC $\sffn(\cdot)$ with $2L$ layers.
                Assume that the input dimension of each even layer of the source FC is $n_{\mathrm{FC}}$, and the input dimension of each odd layer is $d_1$.
                Furthermore, for simplicity, we assume key and value dimensions of $\sattn(\cdot)$ are the same dimension $n_{\mathrm{MHA}}$. 
                Then, with probability at least $1-\epsilon$ for $0 < \epsilon < 1$, there exists a choice of binary masks that satisfies
                \begin{align*}
                    \max_{i \in [T]} \left\|\sblk(\bm{x}_i) - \tblk(\bm{x}_i)\right\| \le \epsilon,
                \end{align*}
                if the hidden dimensions of the source MHA and FC satisfy
                \begin{align*}
                    n_{\mathrm{MHA}} &\ge d_1 C\log\left(\frac{32\alpha^3 H d_1^{\frac{3}{2}}}{\epsilon}\right), \\
                    n_{\mathrm{FC}} &\ge d_1 C\log\left(\frac{24 \alpha L H d_1^{\frac{5}{2}}}{\epsilon}\right),
                \end{align*}
                for some universal constant $C>0$.
                \begin{proof}
                    Our proof strategy is first to apply the SLT existence theorem to the attention mechanism, and then apply the result for FCs.
                    From \Cref{thr:slt_existence_attn_mech_app}, with probability at least $1 - \frac{\epsilon}{4}$, there exists a choice of binary masks so that $\sattn(\cdot)$ satisfies 
                    \begin{align}
                        \label{eq:attn_bound}
                        \|\sattn(\bm{x}_i) - \tattn(\bm{x}_i)\| \le \frac{\epsilon}{4}.
                    \end{align}
                    This inequality \Cref{eq:attn_bound} implies
                    \begin{align*}
                        \|\sattn(\bm{x}_i) - \tattn(\bm{x}_i)\| &\le \frac{\epsilon}{4}, \\
                        \Longrightarrow
                        \|\sattn(\bm{x}_i)\| 
                        &\le 
                        \frac{\epsilon}{4} + \|\tattn(\bm{x}_i)\| 
                        \le
                        \frac{\epsilon}{4} + \alpha H\sqrt{d_1}. 
                    \end{align*}
                    Therefore, the norm of the input vector of the source FC satisfies
                    \begin{align}
                    \nonumber
                        \|\sattn(\bm{x}_i)^\top + \bm{x}\| 
                        &\le
                        \|\sattn(\bm{x}_i)\| + \alpha \\
                        \nonumber
                        &\le
                        \frac{\epsilon}{4} + \alpha (H\sqrt{d_1} + 1) \\
                        \label{eq:fc_input_bound}
                        &\le
                        3\alpha H \sqrt{d_1}.
                    \end{align}
                    Assume that this upper bound of \Cref{eq:fc_input_bound} holds.
                    Now, applying \Cref{thr:slt_existence_within_ffn_app} to the source FC, if 
                    \begin{align*}
                         n_{\mathrm{FC}} 
                         &\ge 
                         d_1 C\log\left(\frac{4 L d_1^2 \cdot 6\alpha H \sqrt{d_1}}{\epsilon}\right) \\
                         &= d_1 C\log\left(\frac{24 \alpha L H d_1^{\frac{5}{2}}}{\epsilon}\right),
                    \end{align*}
                    with probability at least $1 - \frac{\epsilon}{6\alpha H\sqrt{d_1}}$, there exists a choice of binary pruning masks so that $\sffn(\cdot)$ satisfies 
                    \begin{align*}
                        \|\sffn(\sattn(\bm{x}_i)^\top + \bm{x}_i) - &\tffn(\sattn(\bm{x}_i)^\top + \bm{x}_i)\| \\
                        &\le 
                        \exp\left(\frac{3\alpha H \sqrt{d_1}\epsilon}{2\cdot6\alpha H\sqrt{d_1}}\right) - 1 \\
                        &= 
                        \exp\left(\frac{\epsilon}{4}\right) - 1.
                    \end{align*}
                    Finally, we bound the total error between the source and target transformer blocks: 
                    \begin{align*}
                        \max_{i \in [T]}\|\sblk(\bm{x}_i) - \tblk(\bm{x}_i)\| 
                        &= 
                        \max_{i \in [T]}\|\sffn(\sattn(\bm{x}_i)^\top + \bm{x}_i) + \sattn(\bm{x}_i)^\top + \bm{x} - \tffn(\tattn(\bm{x}_i)^\top + \bm{x}_i) - \tattn(\bm{x}_i)^\top - \bm{x}\| \\
                        &\le 
                        \max_{i \in [T]}\Bigl(\|\sffn(\sattn(\bm{x}_i)^\top + \bm{x}_i) - \tffn(\sattn(\bm{x}_i)^\top + \bm{x}_i) \| \\
                        &\quad+ \|\tffn(\sattn(\bm{x}_i)^\top + \bm{x}_i) - \tffn(\tattn(\bm{x}_i)^\top + \bm{x}_i)\| + \|\sattn(\bm{x}_i) - \tattn(\bm{x}_i)\|\Bigr) \\
                        &\le
                        \exp\left(\frac{\epsilon}{4}\right) - 1 + \max_{i \in [T]} 2 \|\sattn(\bm{x}_i) - \tattn(\bm{x}_i)\| \\
                        &\le 
                        \frac{\epsilon}{2} + \frac{\epsilon}{2} \\
                        &= \epsilon.
                    \end{align*}
                    From a union bound, the probability that this approximation holds is at least $1-\epsilon$:
                    \begin{align*}
                        1 - \frac{\epsilon}{4} - \frac{\epsilon}{6\alpha H \sqrt{d_1}} \ge 1 - \epsilon.
                    \end{align*}
                \end{proof}
            \end{theorem}

        \subsection{SLT Existence Within Transformers Without Normalization Layers}
        \label{subsec:theorem_transformer}
        By exploiting the SLT existence theorem for transformer blocks (\Cref{thr:slt_existence_within_attn_block_app}), we prove the SLT existence theorem for transformers without normalization layers.
        
        We firstly prove the two lemmas used in the proof of the theorem.
            \begin{lemma}\label{lem:output_diff_attn_app}
                Let $\bm{X}'=[\bm{x}'_1, \dots, \bm{x}'_T]^\top$ be a perturbed input matrix, which satisfies $\max_{i\in[T]}\|\bm{x}_i - \bm{x}'_i\| \le \epsilon_{\max}$
                Then, an arbitrary target MHA $\tattn(\cdot)$ holds the following inequality:
                \begin{align*}
                    \|\tattn(\bm{x}_i) - &\tattn(\bm{x}'_i)\|
                    \le 
                    H\sqrt{d_1}( \alpha \left( \exp(4\alpha \epsilon_{\max}) - 1 \right) + \epsilon_{\max}).
                \end{align*}
                \begin{proof}
                We begin by analyzing the upper bound of differences for different inputs:
                    \begin{align*}
                        \|\bm{x}_i^\top\bm{W}_{\mathrm{QK}}^{(j)}\bm{X}^\top - \bm{x}_i'^\top\bm{W}_{\mathrm{QK}}^{(j)}\bm{X}'^\top \|_{\max} 
                        &=
                        \max_{k\in [T]}|\bm{x}_i^\top\bm{W}_{\mathrm{QK}}^{(j)}\bm{x}_k - \bm{x}_i'^\top\bm{W}_{\mathrm{QK}}^{(j)}\bm{x}'_k| \\
                        &\le
                        \max_{k\in [T]}\left(|\bm{x}_i^\top\bm{W}_{\mathrm{QK}}^{(j)}\bm{x}_k - \bm{x}_i'^\top\bm{W}_{\mathrm{QK}}^{(j)}\bm{x}_k | + | \bm{x}_i'^\top\bm{W}_{\mathrm{QK}}^{(j)}\bm{x}_k - \bm{x}_i'^\top\bm{W}_{\mathrm{QK}}^{(j)}\bm{x}'_k|\right) \\
                        &\le
                        \max_{k\in [T]}\left(\|\bm{x}_i - \bm{x}'_i\|\|\bm{W}_{\mathrm{QK}}^{(j)}\|\|\bm{x}_k\| + \| \bm{x}'_i\|\|\bm{W}_{\mathrm{QK}}^{(j)}\|\|\bm{x}_k - \bm{x}'_k\|\right) \\
                        &\le
                        \alpha\epsilon_{\max} + \alpha\epsilon_{\max} \\
                        &=
                        2\alpha\epsilon_{\max}.
                    \end{align*}
                    Applying \Cref{lem:softmax_diff_bound_app}, the following inequality holds:
                    \begin{align*}
                        &\|\softmax(\bm{x}_i^\top\bm{W}_{\mathrm{QK}}^{(j)}\bm{X}^\top; \bm{a}_i)\bm{X}\bm{W}_{\mathrm{VO}}^{(j)} - \softmax(\bm{x}_i'^\top\bm{W}_{\mathrm{QK}}^{(j)}\bm{X}'^\top; \bm{a}_i)\bm{X}'\bm{W}_{\mathrm{VO}}^{(j)} \| \\
                        &\le
                        \|\softmax(\bm{x}_i^\top\bm{W}_{\mathrm{QK}}^{(j)}\bm{X}^\top; \bm{a}_i)\bm{X} -  \softmax(\bm{x}_i'^\top\bm{W}_{\mathrm{QK}}^{(j)}\bm{X}'^\top; \bm{a}_i)\bm{X}\| + \| \softmax(\bm{x}_i'^\top\bm{W}_{\mathrm{QK}}^{(j)}\bm{X}'^\top; \bm{a}_i)\bm{X} - \softmax(\bm{x}_i'^\top\bm{W}_{\mathrm{QK}}^{(j)}\bm{X}'^\top; \bm{a}_i)\bm{X}' \| \\
                        &=
                        \|\softmax(\bm{x}_i^\top\bm{W}_{\mathrm{QK}}^{(j)}\bm{X}^\top; \bm{a}_i)\bm{X} -  \softmax(\bm{x}_i'^\top\bm{W}_{\mathrm{QK}}^{(j)}\bm{X}'^\top; \bm{a}_i)\bm{X}\| + \| \softmax(\bm{x}_i'^\top\bm{W}_{\mathrm{QK}}^{(j)}\bm{X}'^\top; \bm{a}_i)(\bm{X} - \bm{X}') \| \\
                        &\le
                        \sqrt{d_1} \alpha \left( \exp(4\alpha \epsilon_{\max}) - 1 \right) + \sqrt{d_1}\epsilon_{\max}. \\
                    \end{align*}
                    Then, we have the following bound:
                    \begin{align*}
                        \|\tattn(\bm{x}_i) - \tattn(\bm{x}')\| 
                        &=
                        \|\sum_{j=1}^{H}\softmax(\bm{x}_i^\top\bm{W}_{\mathrm{QK}}^{(j)}\bm{X}^\top; \bm{a}_i)\bm{X}\bm{W}_{\mathrm{VO}}^{(j)} - \sum_{j=1}^{H}\softmax(\bm{x}_i'^\top\bm{W}_{\mathrm{QK}}^{(j)}\bm{X}'^\top; \bm{a}_i)\bm{X}'\bm{W}_{\mathrm{VO}}^{(j)} \| \\
                        &\le 
                        H\sqrt{d_1}( \alpha \left( \exp(4\alpha \epsilon_{\max}) - 1 \right) + \epsilon_{\max}).
                    \end{align*}
                \end{proof}
            \end{lemma}

             \begin{lemma}\label{lem:output_diff_attn_blk_app}
                An arbitrary target Attention block $\tblk(\cdot)$ holds the following inequality:
                \begin{align*}
                    \|\tblk(\bm{x}_i) - &\tblk(\bm{x}'_i)\|
                    \le H\sqrt{d_1}( \alpha \left( \exp(4\alpha \epsilon_{\max}) - 1 \right) + 2\epsilon_{\max}).
                \end{align*}
                \begin{proof}
                    From \Cref{lem:output_diff_attn_app}, we have the upper bound as follows:
                    \begin{align*}
                        \|\tblk(\bm{x}_i) - \tblk(\bm{x}_i')\|
                        &=
                        \|\tffn(\tattn(\bm{x}_i)^\top + \bm{x}_i) + \tattn(\bm{x}_i)^\top + \bm{x}_i - \tffn(\tattn(\bm{x}_i')^\top + \bm{x}_i') - \tattn(\bm{x}_i')^\top - \bm{x}_i'\| \\
                        &\le
                        \|\tffn(\tattn(\bm{x}_i)^\top + \bm{x}_i) - \tffn(\tattn(\bm{x}_i')^\top + \bm{x}_i') \|  + \|\tattn(\bm{x}_i) - \tattn(\bm{x}_i') \| + \| \bm{x}_i - \bm{x}_i'\| \\
                        &\le
                        2\|\tattn(\bm{x}_i) - \tattn(\bm{x}_i') \| + 2\| \bm{x}_i - \bm{x}_i'\| \\
                        &\le
                        2H\sqrt{d_1}(\alpha(\exp(4\alpha\epsilon_{\max})-1) + \epsilon_{\max}) + 2\epsilon_{\max} \\
                        &\le
                        2H\sqrt{d_1}(\alpha(\exp(4\alpha\epsilon_{\max})-1) + 2\epsilon_{\max})
                    \end{align*}
                \end{proof}
            \end{lemma}
            
            \begin{theorem}
                Assume $B\ge 2$, and let 
                \[
                    \ttrans(\bm{x}_i):=\tblk^{(B)}(\tblk^{(B-1)}\dots \tblk^{(1)}(\bm{x}_i))
                \] 
                be a target transformer with $B$ blocks.
                Let 
                \[
                    \strans(\bm{x}_i):=\sblk^{(B)}(\sblk^{(B-1)}...\sblk^{(1)}(\bm{x}_i))
                \]
                be a pruned random transformer with $B$ layers.
                Then, with probability at least $1 - \epsilon$ for $0 < \epsilon < 1$, there exists a choice of binary masks that satisfies
                \begin{align*}
                    \max_{i \in [T]}\|\strans(\bm{x}_i) - \ttrans(\bm{x}_i)\| \le \epsilon,
                \end{align*}
                if the hidden dimensions of the $b$-th source MHA and FC satisfy
                \begin{align*}
                    n_{\mathrm{MHA}}^{(b)}
                    &\ge
                    d_1 C\log\left( \frac{c_1^{f_1(b, B)} H^{f_2(b, B)} d_1^{f_3(b, B)}}{\epsilon} \right) \\
                    n_{\mathrm{FC}}^{(b)}
                    &\ge
                    d_1 C\log\left( \frac{c_2^{g_1(b, B)} L H^{g_2(b, B)} d_1^{g_3(b, B)}}{\epsilon} \right)
                \end{align*}
                for some universal constant $C>0$ and constants ${c_1, c_2>0}$ including $\alpha$.
                Here, $f_1, f_2, f_3$ and $g_1, g_2, g_3$ are quadratic functions of $B, b$.
                \begin{proof}
                We analyze the approximation errors in each block sequentially and identify the accumulated error in the last block. 
                    \paragraph{Notation for the Proof:}
                    Let $\bm{x}_i^{(b)}$ be an input vector to the $b$-th target block:
                    \begin{align*}
                        \bm{x}_i^{(b)} 
                        = 
                        \begin{dcases*}
                            \bm{x}_i & if $b=1$, \\
                            \tblk^{(b-1)}(\bm{x}_i^{(b-1)}) & if $2\le b \le B$.
                        \end{dcases*}
                    \end{align*}
                    Then, the final output of the target transformer is $\ttrans(\bm{x}_i) = \tblk(\bm{x}_i^{(B)})$.
                    The spectral norm of these input vectors is
                    \begin{align*}
                        \|\bm{x}_i^{(b)}\| 
                        = 
                        \begin{dcases*}
                            \begin{aligned}
                                \|\bm{x}_i\| 
                                &= \alpha \\
                                &=: \beta_1
                            \end{aligned}
                             & if $b=1$, \\
                            \begin{aligned}
                            &\|\tblk^{(b-1)}(\bm{x}_i^{(b-1)})\| \\
                            &\qquad\le 2(H\sqrt{d_1}+1)\|\bm{x}_i^{(b-1)}\| \\
                            &\qquad\le \alpha(2(H\sqrt{d_1}+1))^{b-1} \\
                            &\qquad\le \alpha(4H\sqrt{d_1})^{b-1} \\
                            &\qquad=: 
                            \beta_{b}
                            \end{aligned}
                            & if $2 \le b \le B$.
                        \end{dcases*}
                        \end{align*}
                    Similarly, let $\bm{x}_i'^{(b)}$ be an input vector to the $b$-th source block:
                    \begin{align*}
                        \bm{x}_i'^{(b)} 
                        = 
                        \begin{dcases*}
                            \bm{x}_i & if $b=1$, \\
                            \sblk^{(b-1)}(\bm{x}_i'^{(b-1)}) & if $2\le b \le B$.
                        \end{dcases*}
                    \end{align*}
                    Then, the final output of the source transformer is $\strans(\bm{x}_i) = \sblk(\bm{x}_i'^{(B)})$.
                    \paragraph{First Block Error:}
                        From \Cref{thr:slt_existence_within_attn_block_app}, if
                        \begin{align*}
                            n_{\mathrm{MHA}}^{(1)} 
                            &\ge 
                            d_1 C\log\left(\frac{32\beta_1^3 H d_1^{\frac{3}{2}}2^{B-1}\prod_{j=2}^{B} 16 H\sqrt{d_1}\beta_j^2}{\epsilon}\right), \\
                            n_{\mathrm{FC}}^{(1)} &\ge d_1 C\log\left(\frac{24\beta_1 L H d_1^{\frac{5}{2}}2^{B-1}\prod_{j=2}^{B} 16 H\sqrt{d_1}\beta_j^2}{\epsilon}\right),
                        \end{align*}
                        then with probability at least $1 - \frac{\epsilon}{2^{B-1}\prod_{j=2}^{B} 16 H\sqrt{d_1}\beta_j^2}$, the following inequality holds independently of the input index $i$:
                        \begin{align}
                            \nonumber
                            \|\bm{x}_i^{(2)} - \bm{x}_i'^{(2)}\| 
                            &= 
                            \|\sblk^{(1)}(\bm{x}_i) - \tblk^{(1)}(\bm{x}_i)\| \\
                            \label{eq:first_block_error}
                            &\le 
                            \frac{\epsilon}{2^{B-1}\prod_{j=2}^{B} 16 H\sqrt{d_1}\beta_j^2}.
                        \end{align}
                        This inequality \Cref{eq:first_block_error} implies the upper bound of $\|\bm{x}_i'^{(2)}\|$:
                        \begin{align}
                            \nonumber
                            \|\bm{x}_i^{(2)} - \bm{x}_i'^{(2)}\| 
                            &\le 
                            \frac{\epsilon}{2^{B-1}\prod_{j=2}^{B} 16 H\sqrt{d_1}\beta_j^2} \\
                            \nonumber
                            \Longrightarrow 
                            \|\bm{x}_i'^{(2)}\|
                            &\le
                            \frac{\epsilon}{2^{B-1}\prod_{j=2}^{B} 16 H\sqrt{d_1}\beta_j^2} + \|\bm{x}_i^{(2)}\| \tag{From the triangle inequality.} \\
                            \nonumber
                            &\le
                            \frac{\epsilon}{2^{B-1}\prod_{j=2}^{B} 16 H\sqrt{d_1}\beta_j^2} + \beta_2 \\
                            \label{eq:first_block_bound}
                            &\le
                            2\beta_2.                            
                        \end{align}
                        
                    \paragraph{Second Block Error:}
                        We assume the approximation of the first block is successful (\ie, \Cref{eq:first_block_bound} holds).
                        Then, from \Cref{lem:output_diff_attn_blk_app}, the following bound holds:
                        \begin{align*}
                            &\|\tblk^{(2)}(\bm{x}_i^{(2)}) - \tblk^{(2)}(\bm{x}_i'^{(2)})\| \\
                            &\le
                            H\sqrt{d_1}\left(\beta_2\left(\exp\left(\frac{4\beta_2\epsilon}{2^{B-1}\prod_{j=2}^{B} 16 H\sqrt{d_1}\beta_j^2}\right) - 1\right)  + \frac{2\epsilon}{2^{B-1}\prod_{j=2}^{B} 16 H\sqrt{d_1}\beta_j^2}\right) \\
                            &=
                            H\sqrt{d_1}\left(\beta_2\exp\left(\frac{1}{4 H\sqrt{d_1}\beta_2}\frac{\epsilon}{2^{B-1}\prod_{j=3}^{B} 16 H\sqrt{d_1}\beta_j^2}\right)  - \beta_2 + \frac{1}{8 H\sqrt{d_1}\beta_2}\frac{\epsilon}{2^{B-1}\prod_{j=3}^{B} 16 H\sqrt{d_1}\beta_j^2}\right) \\
                            &\le
                            H\sqrt{d_1}\left(\frac{1}{2H\sqrt{d_1}}\frac{\epsilon}{2^{B-1}\prod_{j=3}^{B} 16 H\sqrt{d_1}\beta_j^2} + \frac{1}{8 H\sqrt{d_1}\beta_2}\frac{\epsilon}{2^{B-1}\prod_{j=3}^{B} 16 H\sqrt{d_1}\beta_j^2}\right) \tag{$\exp(x) \le 2x+1$ if $0\le x \le 1$.} \\
                            &=
                            \frac{1}{2}\frac{\epsilon}{2^{B-1}\prod_{j=3}^{B} 16 H\sqrt{d_1}\beta_j^2} + \frac{1}{8 \beta_2}\frac{\epsilon}{2^{B-1}\prod_{j=3}^{B} 16 H\sqrt{d_1}\beta_j^2} \\
                            &\le
                            \frac{\epsilon}{2^{B-1}\prod_{j=3}^{B} 16 H\sqrt{d_1}\beta_j^2}.
                        \end{align*}
                        From \Cref{thr:slt_existence_within_attn_block_app}, if
                        \begin{align*}
                            n_{\mathrm{MHA}}^{(2)} 
                            &\ge 
                            d_1 C\log\left(\frac{32(2\beta_2)^3 H d_1^{\frac{3}{2}}2^{B-1}\prod_{j=3}^{B} 16 H\sqrt{d_1}\beta_j^2}{\epsilon}\right) \\
                            n_{\mathrm{FC}}^{(2)} 
                            &\ge d_1 C\log\left(\frac{24(2\beta_2) L H d_1^{\frac{5}{2}}2^{B-1}\prod_{j=3}^{B} 16 H\sqrt{d_1}\beta_j^2}{\epsilon}\right),
                        \end{align*}
                        then with probability at least $1 - \frac{\epsilon}{2^{B-1}\prod_{j=3}^{B} 16 H\sqrt{d_1}\beta_j^2}$, the following inequality holds:
                        \begin{align*} 
                            \|\sblk^{(2)}(\bm{x}_i'^{(2)}) - \tblk^{(2)}(\bm{x}_i'^{(2)})\|
                            &\le 
                            \frac{\epsilon}{2^{B-1}\prod_{j=3}^{B} 16 H\sqrt{d_1}\beta_j^2}.
                        \end{align*}  
                        Therefore, we have
                        \begin{align}
                            \nonumber
                            \|\bm{x}_i^{(3)} - \bm{x}_3'^{(3)}\|
                            &=
                            \|\tblk^{(2)}(\bm{x}_i^{(2)}) - \sblk^{(2)}(\bm{x}_i'^{(2)})\| \\
                            \nonumber
                            &=
                            \|\tblk^{(2)}(\bm{x}_i^{(2)}) - \tblk^{(2)}(\bm{x}_i'^{(2)}) + \tblk^{(2)}(\bm{x}_i'^{(2)}) - \sblk^{(2)}(\bm{x}_i'^{(2)})\| \\
                            \nonumber
                            &\le
                            \|\tblk^{(2)}(\bm{x}_i^{(2)}) - \tblk^{(2)}(\bm{x}_i'^{(2)}) \| + \| \tblk^{(2)}(\bm{x}_i'^{(2)}) - \sblk^{(2)}(\bm{x}_i'^{(2)})\| \\
                            \nonumber
                            &\le
                            \frac{\epsilon}{2^{B-1}\prod_{j=3}^{B} 16 H\sqrt{d_1}\beta_j^2} + \frac{\epsilon}{2^{B-1}\prod_{j=3}^{B} 16 H\sqrt{d_1}\beta_j^2} \\
                            \label{eq:second_block_error}
                            &=
                            \frac{\epsilon}{2^{B-2}\prod_{j=3}^{B} 16 H\sqrt{d_1}\beta_j^2}.
                        \end{align}
                        This inequality \Cref{eq:second_block_error} implies the upper bound of $\|\bm{x}_i'^{(3)}\|$:
                        \begin{align}
                            \nonumber
                            \|\bm{x}_i^{(3)} - \bm{x}_i'^{(3)}\| 
                            &\le 
                            \frac{\epsilon}{2^{B-2}\prod_{j=3}^{B} 16 H\sqrt{d_1}\beta_j^2} \\
                            \nonumber
                            \Longrightarrow 
                            \|\bm{x}_i'^{(3)}\|
                            &\le
                            \frac{\epsilon}{2^{B-2}\prod_{j=3}^{B} 16 H\sqrt{d_1}\beta_j^2} + \|\bm{x}_i^{(3)}\| \tag{From the triangle inequality.} \\
                            \nonumber
                            &\le
                            \frac{\epsilon}{2^{B-2}\prod_{j=3}^{B} 16 H\sqrt{d_1}\beta_j^2} + \beta_3 \\
                            \label{eq:second_block_bound}
                            &\le
                            2\beta_3.                            
                        \end{align}
                        
                    \paragraph{Third Block Error:}
                        We assume the approximation of second block is succeessful (\ie, \Cref{eq:second_block_bound} holds).
                        Then, from \Cref{lem:output_diff_attn_blk_app}, the following bound holds:
                        \begin{align*}
                            &\|\tblk^{(3)}(\bm{x}_i^{(3)}) - \tblk^{(3)}(\bm{x}_i'^{(3)})\| \\
                            &\le
                            H\sqrt{d_1}\left(\beta_3\left(\exp\left(4\beta_3\frac{\epsilon}{2^{B-2}\prod_{j=3}^{B} 16 H\sqrt{d_1}\beta_j^2}\right) - 1\right) + 2\frac{\epsilon}{2^{B-2}\prod_{j=3}^{B} 16 H\sqrt{d_1}\beta_j^2}\right) \\
                            &=
                            H\sqrt{d_1}\left(\beta_3\left(\exp\left(\frac{1}{4 H\sqrt{d_1}\beta_3}\frac{\epsilon}{2^{B-2}\prod_{j=4}^{B} 16 H\sqrt{d_1}\beta_j^2}\right) - 1\right) + \frac{1}{8 H\sqrt{d_1}\beta_3}\frac{\epsilon}{2^{B-2}\prod_{j=4}^{B} 16 H\sqrt{d_1}\beta_j^2}\right) \\
                            &\le
                            H\sqrt{d_1}\left(\frac{1}{2H\sqrt{d_1}}\frac{\epsilon}{2^{B-2}\prod_{j=4}^{B} 16 H\sqrt{d_1}\beta_j^2} + \frac{1}{8 H\sqrt{d_1}\beta_3}\frac{\epsilon}{2^{B-2}\prod_{j=4}^{B} 16 H\sqrt{d_1}\beta_j^2}\right) \tag{$\exp(x) \le 2x+1$ if $0\le x \le 1$.} \\
                            &=
                            \frac{1}{2}\frac{\epsilon}{2^{B-2}\prod_{j=4}^{B} 16 H\sqrt{d_1}\beta_j^2} + \frac{1}{8 \beta_3}\frac{\epsilon}{2^{B-2}\prod_{j=4}^{B} 16 H\sqrt{d_1}\beta_j^2} \\
                            &\le
                            \frac{\epsilon}{2^{B-2}\prod_{j=4}^{B} 16 H\sqrt{d_1}\beta_j^2}.
                        \end{align*}
                        From \Cref{thr:slt_existence_within_attn_block_app}, if
                        \begin{align*}
                            n_{\mathrm{MHA}}^{(3)} 
                            &\ge 
                            d_1 C\log\left(\frac{32(2\beta_3)^3 H d_1^{\frac{3}{2}}2^{B-2}\prod_{j=4}^{B} 16 H\sqrt{d_1}\beta_j^2}{\epsilon}\right) \\
                            n_{\mathrm{FC}}^{(3)} 
                            &\ge 
                            d_1 C\log\left(\frac{24(2\beta_3) L H d_1^{\frac{5}{2}}2^{B-2}\prod_{j=4}^{B} 16 H\sqrt{d_1}\beta_j^2}{\epsilon}\right),
                        \end{align*}
                        then with probability at least $1 - \frac{\epsilon}{2^{B-2}\prod_{j=4}^{B} 16 H\sqrt{d_1}\beta_j^2}$, the following inequality holds:
                        \begin{align*} 
                            \|\sblk^{(3)}(\bm{x}_i'^{(3)}) - \tblk^{(3)}(\bm{x}_i'^{(3)})\|
                            &\le 
                            \frac{\epsilon}{2^{B-2}\prod_{j=4}^{B} 16 H\sqrt{d_1}\beta_j^2}.
                        \end{align*}  
                        Therefore, we have 
                        \begin{align}
                            \nonumber
                            &\|\bm{x}_i^{(4)} - \bm{x}_3'^{(4)}\| \\
                            &=
                            \|\tblk^{(3)}(\bm{x}_i^{(3)}) - \sblk^{(3)}(\bm{x}_i'^{(3)})\| \\
                            \nonumber
                            &=
                            \|\tblk^{(3)}(\bm{x}_i^{(3)}) - \tblk^{(3)}(\bm{x}_i'^{(3)}) + \tblk^{(3)}(\bm{x}_i'^{(3)}) - \sblk^{(3)}(\bm{x}_i'^{(3)})\| \\
                            \nonumber
                            &\le
                            \|\tblk^{(3)}(\bm{x}_i^{(3)}) - \tblk^{(3)}(\bm{x}_i'^{(3)}) \|  + \| \tblk^{(3)}(\bm{x}_i'^{(3)}) - \sblk^{(3)}(\bm{x}_i'^{(3)})\| \\
                            \nonumber
                            &\le
                            \frac{\epsilon}{2^{B-2}\prod_{j=4}^{B} 16 H\sqrt{d_1}\beta_j^2} + \frac{\epsilon}{2^{B-2}\prod_{j=4}^{B} 16 H\sqrt{d_1}\beta_j^2} \\
                            \label{eq:third_block_error}
                            &=
                            \frac{\epsilon}{2^{B-3}\prod_{j=4}^{B} 16 H\sqrt{d_1}\beta_j^2}.
                        \end{align}
                        This inequality \Cref{eq:third_block_error} implies the upper bound of $\|\bm{x}_i'^{(4)}\|$:
                        \begin{align}
                            \nonumber
                            \|\bm{x}_i^{(4)} - \bm{x}_i'^{(4)}\| 
                            &\le 
                            \frac{\epsilon}{2^{B-3}\prod_{j=4}^{B} 16 H\sqrt{d_1}\beta_j^2} \\
                            \nonumber
                            \Longrightarrow 
                            \|\bm{x}_i'^{(4)}\|
                            &\le
                            \frac{\epsilon}{2^{B-3}\prod_{j=4}^{B} 16 H\sqrt{d_1}\beta_j^2} + \|\bm{x}_i^{(4)}\| \tag{From the triangle inequality.} \\
                            \nonumber
                            &\le
                            \frac{\epsilon}{2^{B-3}\prod_{j=4}^{B} 16 H\sqrt{d_1}\beta_j^2} + \beta_4 \\
                            \label{eq:third_block_bound}
                            &\le
                            2\beta_4.                            
                        \end{align}
                        
                    \paragraph{$(B-1)$-th Block Error:}
                        By repeating the same proof procedure as above in each block, we can propagate the error to $(B-1)$-th block.
                        We assume all first-to-$(B-2)$-th block approximations are successful.
                        Then, from \Cref{lem:output_diff_attn_blk_app}, the following bound holds:
                        \begin{align*}
                            &\|\tblk^{(B-1)}(\bm{x}_i^{(B-1)}) - \tblk^{(B-1)}(\bm{x}_i'^{(B-1)})\| \\
                            &\le
                            H\sqrt{d_1}\left(\beta_{B-1}\left(\exp\left(4\beta_{B-1}\frac{\epsilon}{2^2\prod_{j=B-1}^{B} 16 H\sqrt{d_1}\beta_j^2}\right) - 1 \right) + 2\frac{\epsilon}{2^2\prod_{j=B-1}^{B} 16 H\sqrt{d_1}\beta_j^2}\right) \\
                            &=
                            H\sqrt{d_1}\left(\beta_{B-1}\left(\exp\left(\frac{1}{4 H\sqrt{d_1}\beta_{B-1}}\frac{\epsilon}{2^2 \cdot 16 H\sqrt{d_1}\beta_B^2}\right) - 1 \right) + \frac{1}{8 H\sqrt{d_1}\beta_{B-1}}\frac{\epsilon}{2^2 \cdot 16 H\sqrt{d_1}\beta_B^2}\right) \\
                            &\le
                            H\sqrt{d_1}\left(\frac{1}{2H\sqrt{d_1}}\frac{\epsilon}{2^2 \cdot 16 H\sqrt{d_1}\beta_B^2} + \frac{1}{8 H\sqrt{d_1}\beta_{B-1}}\frac{\epsilon}{2^2 \cdot 16 H\sqrt{d_1}\beta_B^2}\right) \tag{$\exp(x) \le 2x+1$ if $0\le x \le 1$.} \\
                            &=
                            \frac{1}{2}\frac{\epsilon}{2^2 \cdot 16 H\sqrt{d_1}\beta_B^2} + \frac{1}{8 \beta_{B-1}}\frac{\epsilon}{2^2 \cdot 16 H\sqrt{d_1}\beta_B^2} \\
                            &\le
                            \frac{\epsilon}{2^2\cdot 16 H\sqrt{d_1}\beta_B^2}.
                        \end{align*}
                        From \Cref{thr:slt_existence_within_attn_block_app}, if
                        \begin{align*}
                            n_{\mathrm{MHA}}^{(B-1)} 
                            &\ge 
                            d_1 C\log\left(\frac{32(2\beta_{B-1})^3 H d_1^{\frac{3}{2}}2^2\cdot 16 H\sqrt{d_1}\beta_B^2}{\epsilon}\right) \\
                            n_{\mathrm{FC}}^{(B-1)} 
                            &\ge 
                            d_1 C\log\left(\frac{24(2\beta_{B-1}) L H d_1^{\frac{5}{2}}2^2\cdot 16 H\sqrt{d_1}\beta_B^2}{\epsilon}\right),
                        \end{align*}
                        then with probability at least $1 - \frac{\epsilon}{2^2\cdot 16 H\sqrt{d_1}\beta_B^2}$, the following inequality holds independently of the input index $i$:
                        \begin{align*} 
                            \|\sblk^{(B-1)}(\bm{x}_i'^{(B-1)}) - \tblk^{(B-1)}(\bm{x}_i'^{(B-1)})\| 
                            \le 
                            \frac{\epsilon}{2^2\cdot 16 H\sqrt{d_1}\beta_j^2}.
                        \end{align*}  
                        Therefore, we have the following inequality:
                        \begin{align}
                            \nonumber
                            \|\bm{x}_i^{(B)} - \bm{x}_3'^{(B)}\| 
                            &=
                            \|\tblk^{(B-1)}(\bm{x}_i^{(B-1)}) - \sblk^{(B-1)}(\bm{x}_i'^{(B-1)})\| \\
                            \nonumber
                            &=
                            \|\tblk^{(B-1)}(\bm{x}_i^{(B-1)}) - \tblk^{(B-1)}(\bm{x}_i'^{(B-1)})  + \tblk^{(B-1)}(\bm{x}_i'^{(B-1)}) - \sblk^{(B-1)}(\bm{x}_i'^{(B-1)})\| \\
                            \nonumber
                            &\le
                            \|\tblk^{(B-1)}(\bm{x}_i^{(B-1)}) - \tblk^{(B-1)}(\bm{x}_i'^{(B-1)}) \| + \| \tblk^{(B-1)}(\bm{x}_i'^{(B-1)}) - \sblk^{(B-1)}(\bm{x}_i'^{(B-1)})\| \\
                            \nonumber
                            &\le
                            \frac{\epsilon}{2^2\cdot 16 H\sqrt{d_1}\beta_B^2} + \frac{\epsilon}{2^2\cdot 16 H\sqrt{d_1}\beta_B^2} \\
                            \label{eq:bth_block_error}
                            &=
                            \frac{\epsilon}{2 \cdot 16 H\sqrt{d_1}\beta_B^2}.
                        \end{align}
                        This inequality \Cref{eq:bth_block_error} implies the upper bound of $\|\bm{x}_i'^{(B)}\|$:
                        \begin{align}
                            \nonumber
                            \|\bm{x}_i^{(B)} - \bm{x}_i'^{(B)}\| 
                            &\le 
                            \frac{\epsilon}{2\cdot 16 H\sqrt{d_1}\beta_B^2} \\
                            \nonumber
                            \Longrightarrow 
                            \|\bm{x}_i'^{(B)}\|
                            &\le
                            \frac{\epsilon}{2\cdot 16 H\sqrt{d_1}\beta_B^2} + \|\bm{x}_i^{(B)}\| \tag{From the triangle inequality.} \\
                            \nonumber
                            &\le
                            \frac{\epsilon}{2\cdot 16 H\sqrt{d_1}\beta_B^2} + \beta_B \\
                            \label{eq:bth_block_bound}
                            &\le
                            2\beta_B.                            
                        \end{align}

                    \paragraph{Final Block Error:}
                        We assume all first-to $(B-1)$-th block approximations are successful.
                        Then, from \Cref{lem:output_diff_attn_blk_app}, the following bound holds:
                        \begin{align*}
                            &\|\tblk^{(B)}(\bm{x}_i^{(B)}) - \tblk^{(B)}(\bm{x}_i'^{(B)})\| \\
                            &\le
                            H\sqrt{d_1}\left(\beta_{B}\left(\exp\left(4\beta_{B}\frac{\epsilon}{2\cdot 16 H\sqrt{d_1}\beta_B^2}\right) - 1\right) + 2\frac{\epsilon}{2\cdot 16 H\sqrt{d_1}\beta_B^2}\right) \\
                            &=
                            H\sqrt{d_1}\left(\beta_{B}\left(\exp\left(\frac{1}{8 H\sqrt{d_1}\beta_{B}}\epsilon\right) - 1\right) + \frac{1}{16 H\sqrt{d_1}\beta_{B}}\epsilon \right) \\
                            &\le
                            H\sqrt{d_1}\left(\frac{1}{4H\sqrt{d_1}}\epsilon + \frac{1}{16 H\sqrt{d_1}\beta_{B}}\epsilon\right) \tag{$\exp(x) \le 2x+1$ if $0\le x \le 1$.} \\
                            &=
                            \frac{1}{4}\epsilon + \frac{1}{16 \beta_{B}}\epsilon \\
                            &\le
                            \frac{\epsilon}{2}.
                        \end{align*}
                        From \Cref{thr:slt_existence_within_attn_block_app}, if
                        \begin{align*}
                            n_{\mathrm{MHA}}^{(B)} 
                            &\ge 
                            d_1 C\log\left(\frac{32(2\beta_{B})^3 H d_1^{\frac{3}{2}} \cdot 2}{\epsilon}\right), \\
                            n_{\mathrm{FC}}^{(B)} &\ge d_1 C\log\left(\frac{24(2\beta_{B}) L H d_1^{\frac{5}{2}} \cdot 2}{\epsilon}\right),
                        \end{align*}
                        then with probability at least $1 - \frac{\epsilon}{2}$, the following inequality holds independently of the input index $i$:
                        \begin{align*} 
                            \|\sblk^{(B)}(\bm{x}_i'^{(B)}) - \tblk^{(B)}(\bm{x}_i'^{(B)})\|
                            &\le 
                            \frac{\epsilon}{2}.
                        \end{align*}  
                        Therefore, we finally obtain the following inequality:
                        \begin{align*}
                            \|\ttrans(\bm{x}_i) - \strans(\bm{x}_i)\| 
                            &=
                            \|\tblk^{(B)}(\bm{x}_i^{(B)}) - \sblk^{(B)}(\bm{x}_i'^{(B)})\| \\
                            &=
                            \|\tblk^{(B)}(\bm{x}_i^{(B)}) - \tblk^{(B)}(\bm{x}_i'^{(B)}) + \tblk^{(B)}(\bm{x}_i'^{(B)}) - \sblk^{(B)}(\bm{x}_i'^{(B)})\| \\
                            &\le
                            \|\tblk^{(B)}(\bm{x}_i^{(B)}) - \tblk^{(B)}(\bm{x}_i'^{(B)}) \| + \| \tblk^{(B)}(\bm{x}_i'^{(B)}) - \sblk^{(B)}(\bm{x}_i'^{(B)})\| \\
                            &\le
                            \frac{\epsilon}{2} + \frac{\epsilon}{2} \\
                            &=
                            \epsilon.
                        \end{align*}
                    \paragraph{Success Probability of the Approximation:}
                        By the union bound, the probability that $\|\ttrans(\bm{x}_i) - \strans(\bm{x}_i)\| \le \epsilon$ holds is at least $1-\epsilon$:
                        \begin{align*}
                            &1 - \frac{\epsilon}{2^{B-1}\prod_{j=2}^{B} 16 H\sqrt{d_1}\beta_j^2} - \frac{\epsilon}{2^{B-1}\prod_{j=3}^{B} 16 H\sqrt{d_1}\beta_j^2} - \frac{\epsilon}{2^{B-2}\prod_{j=4}^{B} 16 H\sqrt{d_1}\beta_j^2} - \dots - \frac{\epsilon}{2^{2}\cdot 16 H\sqrt{d_1}\beta_j^2} - \frac{\epsilon}{2} \\
                            &=
                            1 - \frac{2 + \sum_{k=2}^B \prod_{j=2}^k32H\sqrt{d_1}\beta_j^2}{2\prod_{j=2}^B 32H\sqrt{d_1}\beta_j^2}\epsilon \\
                            &=
                            1 - \frac{1}{2\prod_{j=2}^B 32H\sqrt{d_1}\beta_j^2}\epsilon  - \frac{1 + \sum_{k=2}^B \prod_{j=2}^k32H\sqrt{d_1}\beta_j^2}{2\prod_{j=2}^B 32H\sqrt{d_1}\beta_j^2}\epsilon \\
                            &\ge
                            1 - \frac{1}{64}\epsilon - \frac{1 + \sum_{k=2}^B \prod_{j=2}^k32H\sqrt{d_1}\beta_j^2}{2\prod_{j=2}^B 32H\sqrt{d_1}\beta_j^2}\epsilon \\
                            &=
                            1 - \frac{1}{64}\epsilon  - \frac{1 + \sum_{k=2}^B (32H\alpha^2 \sqrt{d_1})^{k-1}\prod_{j=2}^k(16H^2 d_1)^{j-1}}{2(32H\alpha^2\sqrt{d_1})^{B-1}\prod_{j=2}^k(16H^2 d_1)^{j-1}}\epsilon \\   
                            &=
                            1 - \frac{1}{64}\epsilon - \frac{1 + \sum_{k=2}^B (32H\alpha^2 \sqrt{d_1} (16H^2 d_1)^{\frac{1}{2}k})^{k-1}}{2(32H\alpha^2\sqrt{d_1})^{B-1}(16H^2 d_1)^{\frac{1}{2}B(B-1)}}\epsilon \\
                            &\ge
                            1 - \frac{1}{64}\epsilon - \frac{\sum_{k=1}^B (32H\alpha^2 \sqrt{d_1} (16H^2 d_1)^{\frac{1}{2}B})^{k-1}}{2(32H\alpha^2\sqrt{d_1})^{B-1}(16H^2 d_1)^{\frac{1}{2}B(B-1)}}\epsilon \\
                            &\ge
                            1 - \frac{1}{64}\epsilon  - \frac{(32H\alpha^2 \sqrt{d_1} (16H^2 d_1)^{\frac{1}{2}B})^B - 1}{(32H\alpha^2 \sqrt{d_1} (16H^2 d_1)^{\frac{1}{2}B} - 1)}  \cdot \frac{1}{2(32H\alpha^2\sqrt{d_1})^{B-1}(16H^2 d_1)^{\frac{1}{2}B(B-1)}}\epsilon \\
                            &\ge
                            1 - \frac{1}{64}\epsilon  - \frac{(32H\alpha^2 \sqrt{d_1} (16H^2 d_1)^{\frac{1}{2}B})^B}{(32H\alpha^2 \sqrt{d_1} (16H^2 d_1)^{\frac{1}{2}B} - 1)}  \cdot\frac{1}{2(32H\alpha^2\sqrt{d_1})^{B-1}(16H^2 d_1)^{\frac{1}{2}B(B-1)}}\epsilon \\
                            &=
                            1 - \frac{1}{64}\epsilon - \frac{1}{2\left(1 - \frac{1}{32H\alpha^2\sqrt{d_1}(16H^2 d_1)^{\frac{1}{2}B}}\right)}\epsilon \\
                            &\ge
                            1 - \frac{1}{64}\epsilon - \frac{1}{2\left(1 - \frac{1}{32}\right)}\epsilon \\
                            &=
                            1 - \frac{1}{64}\epsilon - \frac{16}{32}\epsilon \\
                            &=
                            1 - \frac{1055}{1984}\epsilon \\
                            &\ge
                            1 - \epsilon.
                        \end{align*}
                    
                \end{proof}
            \end{theorem}

\section{Experimental Details}\label{app:experimental-details}
This section describes the detailed experimental settings.
All experiments can be verified with four NVIDIA H100 SXM5 94GB GPUs.

\subsection{Synthetic Data Experiment}
We construct a synthetic dataset for angular velocity estimation, where each input sequence consists of $T$ two-dimensional vectors $x_1,..., x_T$ such that $x_t = (\cos(\omega t + \theta_0), \sin(\omega t  + \theta_0))$ for some angular velocity $\omega \in [-\pi, \pi]$ and initial phase $\theta_0 \in [0, \pi]$. 
The task is to estimate $\omega$ given the full sequence. 
Each sequence includes a special regression token---similar to the CLS token in BERT~\citep{devlin2019bert}---at the beginning, and the model is trained to predict angular velocity by the regression token initialized to zero.
We generate $10,000$ samples each for training, validation, and test sets, and input sequence lengths vary from $4$ to $256$ during training.
We experiment with MHAs and transformers.
In the MHA experiment, both the source and target MHAs are configured as single-head attention modules, with input and output dimensions of $2$ and $1$, respectively. 
The networks are trained using the AdamW optimizer~\citep{loshchilov2018decoupled} with a batch size of $1024$ and a learning rate of $0.1$. 
Each target MHA is trained for $25$ epochs with weight decay set to $0.01$.
In the transformer experiment, both the source and target models follow the construction described in \Cref{subsec:slt_existence_transformer}. 
Each MHA has a single attention head, and both its input and output dimensions are set to $2$. 
The same regression token is used for both the source and target models to ensure that the approximation quality reflects differences in the behavior of the models rather than token-level discrepancies. 
The query and key dimensions of the target models are set to $8$. 
Target networks are initialized according to the assumptions of our theoretical results. Specifically, entries of the query and key projection matrices are drawn i.i.d. from $U[-n_{\textrm{K}}^{1/4}, n_{\textrm{K}}^{1/4}]$, and those of the value and output projection matrices from $U[-1, 1]$. 
The weights in fully-connected networks are also initialized with $U[-1, 1]$.
Source networks are initialized with Xavier uniform distribution~\citep{pmlr-v9-glorot10a}.
To identify SLTs, we use the weight approximation method in \Cref{lem:weight_approximation}, based on the subset-sum approximation technique of \citet{pensia2020optimal}. 
For each target network, we generate $100$ source networks with random initialization and solve the associated subset-sum problem using Gurobi’s mixed-integer programming solver~\citep{gurobi}. 
In the experiments varying the hidden dimension, the input length is fixed at 4.
We report the mean and standard deviation of the approximation error over these $100$ candidates. 
We also fit exponential decay curves to the approximation error using SciPy~\citep{2020SciPy-NMeth}.

\subsection{Language Modeling Experiment}
We further evaluate our theoretical framework in a practical language modeling setting. 
Here, we search for SLTs using the edge-popup algorithm~\citep{ramanujan2020s}, which searches for accurate subnetworks by assigning scores to each connection and retaining only the top-$k \%$ entries during training.
We set this $k$ as $30$.
We train models from the GPT-2 family~\citep{radford2019language,wolf-etal-2020-transformers} on the WikiText-103 dataset~\citep{merity2017pointer}, using a maximum sequence length of $1024$. 
The weights of these models are initialized based on the GPT-2 initialization scheme: 
they are drawn i.i.d. from a normal distribution with mean $0$ and standard deviation $0.02$.
For the output projection in MHAs and the second layer of the fully-connected ReLU networks, the standard deviation is further scaled by $(2b)^{-1/2}$, where $b$ is the number of transformer blocks.
We train the models for $50$ epochs, with $227$ steps per epoch. 
The AdamW optimizer is used with an initial learning rate of $0.0001$, which is decayed to $0.00001$ via a cosine annealing scheduler~\citep{loshchilov2017sgdr}. 
A linear learning rate warm-up is applied during the first epoch.
For each model size, we repeat training with three different random seeds and report the mean and standard deviation of the best performance.

\end{document}